\documentclass[sigconf]{acmart}
\renewcommand\footnotetextcopyrightpermission[1]{}
\setcopyright{none}
\authorsaddresses{}
\acmConference{}{}{}
\acmBooktitle{}
\acmYear{}

\usepackage{booktabs}
\usepackage[ruled]{algorithm2e} 

\SetAlFnt{\small}
\SetAlCapFnt{\small}
\SetAlCapNameFnt{\small}
\SetAlCapHSkip{0pt}

\usepackage[utf8]{inputenc} 
\usepackage[T1]{fontenc}    
\usepackage{url}            
\usepackage{amsfonts}       
\usepackage{nicefrac}       
\usepackage{microtype}      
\usepackage{listings}
\usepackage{xcolor}         
\usepackage{xspace}
\usepackage{graphicx}
\usepackage{tabularx}
\usepackage{multirow}
\usepackage{float}
\usepackage{makecell}
\usepackage{amsmath}
\usepackage{wrapfig}
\usepackage{bm}
\usepackage{subcaption}
\usepackage{algpseudocode}
\usepackage[capitalize,noabbrev]{cleveref}

\newcommand{\mypara}[1]{\vspace{0pt}\noindent\textbf{#1}}

\lstdefinelanguage{json}{
    basicstyle=\ttfamily\footnotesize,
    numbers=left,
    numberstyle=\tiny,
    stepnumber=1,
    numbersep=8pt,
    showstringspaces=false,
    breaklines=true,
    breakatwhitespace=false,
    frame=single,
    tabsize=2,
    stringstyle=\color{black},
}

\begin{document}

\title{NuiWorld: Exploring a Scalable Framework for End-to-End Controllable World Generation}

\author{Han-Hung Lee}
\affiliation{
 \institution{Simon Fraser University}
 \country{Canada}}
\email{hla300@sfu.ca}

\author{Cheng-Yu Yang}
\affiliation{%
 \institution{National Yang Ming Chiao Tung University}
 \country{Taiwan}
}
\email{cyyang@nycu.edu.tw}

\author{Yu-Lun Liu}
\affiliation{%
 \institution{National Yang Ming Chiao Tung University}
 \country{Taiwan}
}
\email{yulunliu@cs.nycu.edu.tw}

\author{Angel X. Chang}
\affiliation{
 \institution{Simon Fraser University, CIFAR AI Chair, Amii}
 \country{Canada}}
\email{angelx@sfu.ca}

\begin{abstract}
    World generation is a fundamental capability for applications like video games, simulation, and robotics. However, existing approaches face three main obstacles: controllability, scalability, and efficiency. End-to-end scene generation models have been limited by data scarcity. While object-centric generation approaches rely on fixed resolution representations, degrading fidelity for larger scenes. Training-free approaches, while flexible, are often slow and computationally expensive at inference time. We present NuiWorld, a framework that attempts to address these challenges. To overcome data scarcity, we propose a generative bootstrapping strategy that starts from a few input images. Leveraging recent 3D reconstruction and expandable scene generation techniques, we synthesize scenes of varying sizes and layouts, producing enough data to train an end-to-end model. Furthermore, our framework enables controllability through pseudo sketch labels, and demonstrates a degree of generalization to previously unseen sketches. Our approach represents scenes as a collection of variable scene chunks, which are compressed into a flattened vector-set representation. This significantly reduces the token length for large scenes, enabling consistent geometric fidelity across scenes sizes while improving training and inference efficiency.
\end{abstract}

\begin{teaserfigure}
  \centering
  \includegraphics[width=\textwidth]{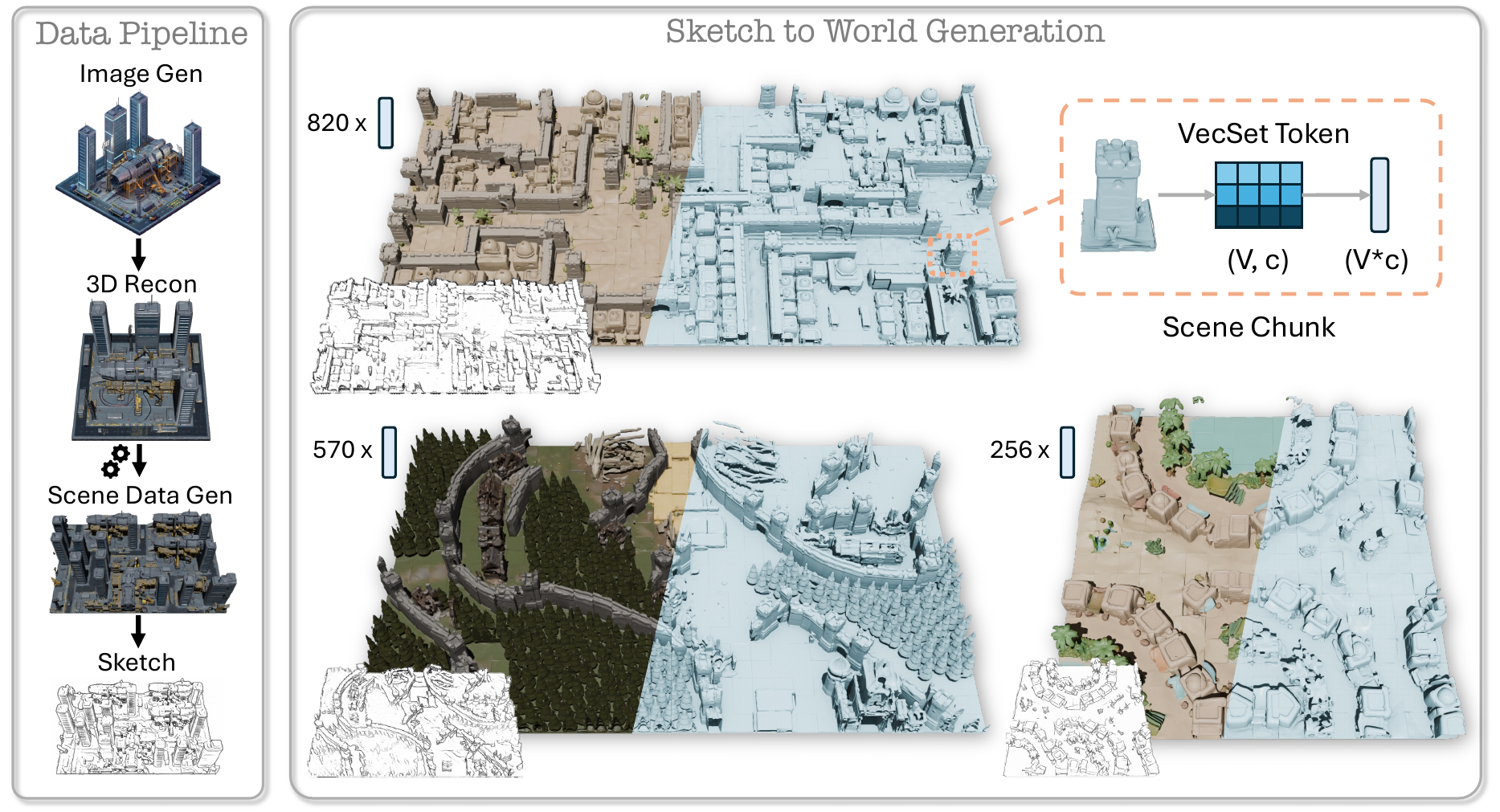}
  \caption{NuiWorld proposes a data pipeline for constructing \textbf{open-domain}, \textbf{variable-sized} scenes with pseudo sketch pairs to train controllable world generation models. Furthermore, our representation scales with scene size while maintaining fidelity across scenes of varying sizes.}
  \label{fig:teaser}
\end{teaserfigure}

\maketitle

\pagestyle{plain}

\section{Introduction}

The ability to generate virtual worlds at a click of a button has wide-ranging applications. For artists, it could enable rapid visualization of environments for concept design. In robotics, generating diverse environments for training can improve generalization. Similar benefits can extend to VR, film and other industries. As a result, fast and open-domain world generation is increasingly important, and we believe will be best achieved through end-to-end models. In this paper, we define world generation as the ability to produce large-scale, open-domain scenes. We argue that addressing the two core challenges of scalability and data scarcity is essential to achieving this and enabling world generation for broader applicability.

\begin{figure}
\centering
\includegraphics[width=\linewidth]{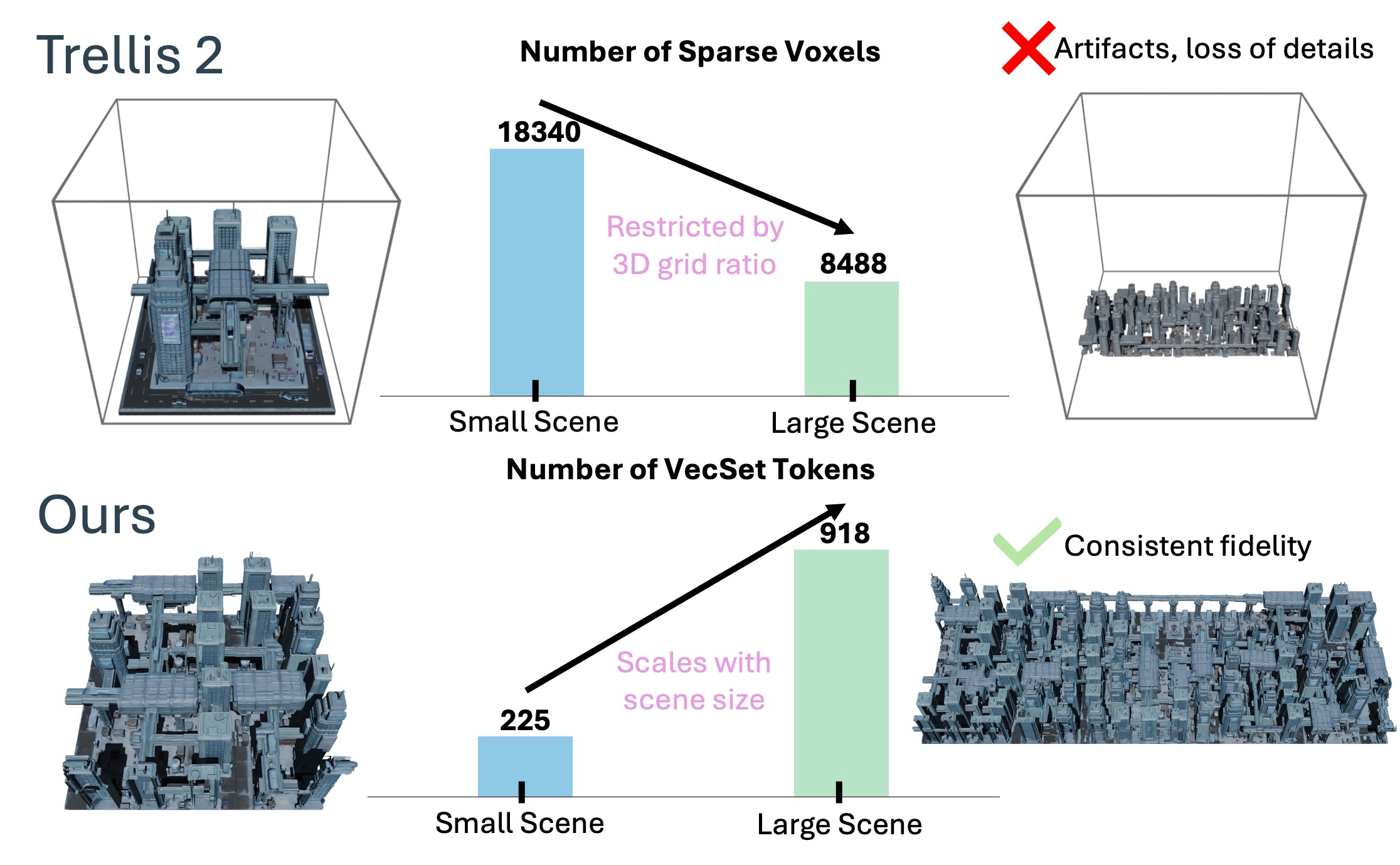}
\vspace{-20pt}
\caption{Object generators like Trellis 2 do not adapt to scene scale or aspect ratio, causing such scenes to be represented with less voxels and degraded fidelity while our method scales with scene sizes and maintains fidelity.}
\label{fig:trellis2_limitation}
\vspace{-20pt}
\end{figure}

\begin{figure*}
\centering
\vspace{-1em}
\includegraphics[width=\linewidth]{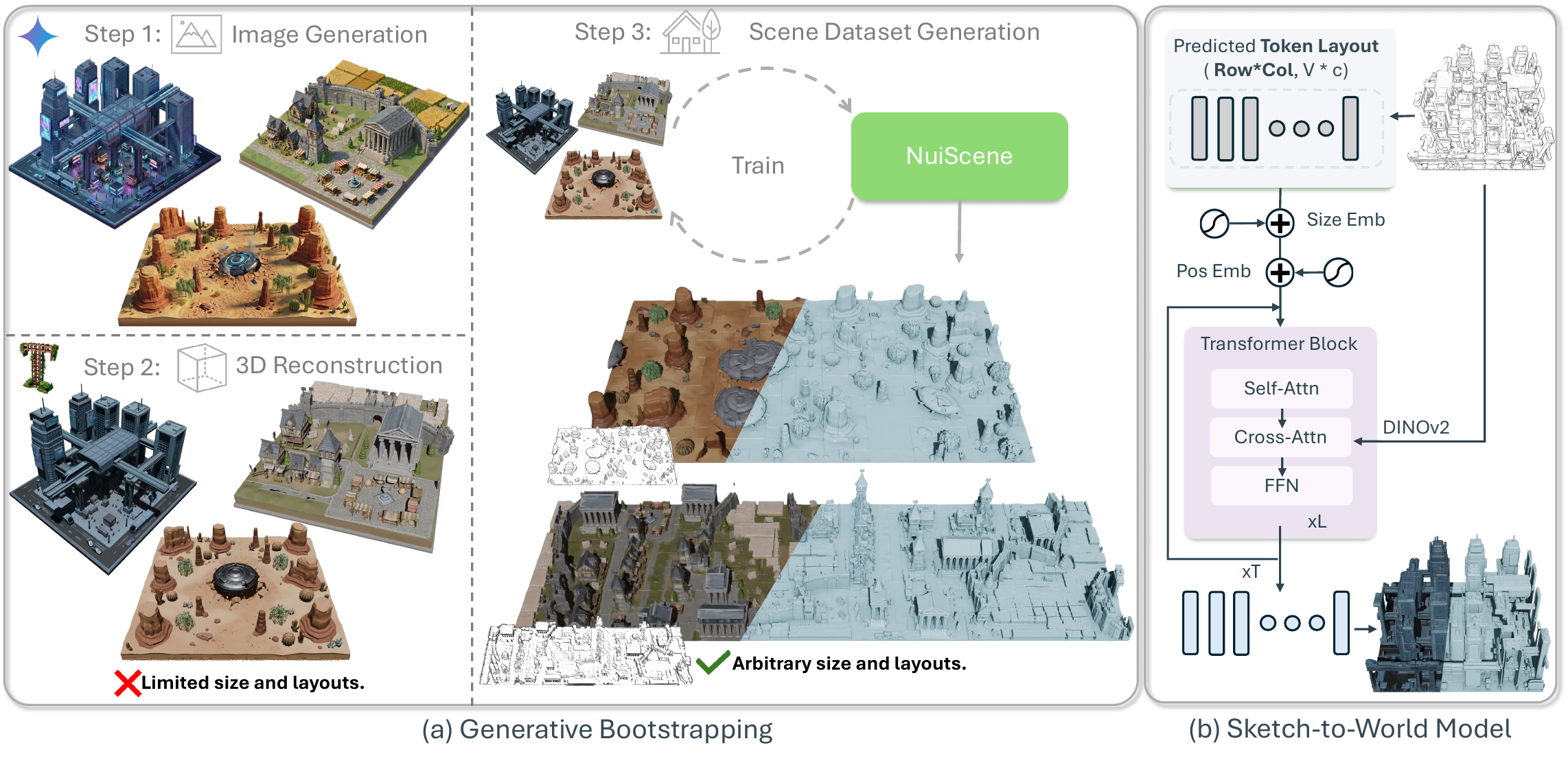}
\caption{Our framework begins with generative bootstrapping, shown on the left. Using Nano Banana to generate images and Trellis 2 to reconstruct 3D scenes. NuiScene is then trained on these scenes to produce new scenes with varying size and layouts. Finally, using these scenes and their pseudo sketches we train our variable-length sketch-to-world model on the right.}
\label{fig:pipeline}
\end{figure*}

With the rise of LLMs and agent-based systems used to tackle tasks with limited training data, these models have also been explored for large-scale scene generation. Several works~\cite{wang2025worldgen, lu2025yo, huang2025majutsucity, wang2025raisecity} employ LLMs to aid the scene generation process. However, this reliance introduces additional latency time and potential costs. Additionally, it remains unclear whether LLMs can reliably perform spatial reasoning for scene layout generation or generalize to open-domain scene layouts. Our approach instead focuses on synthesizing a sufficiently large training dataset of scenes for world generation as seen in~\cref{fig:pipeline}. We propose a generative bootstrapping pipeline that begins with a small set of scenario images produced using text-to-image models. These images are reconstructed into 3D scenes with Trellis 2~\cite{xiang2025native}, which are used to train NuiScene~\cite{Lee_2025_ICCV} to generate varying scene layouts and scales. These scenes are converted into pseudo sketches and used to train a controllable world generation model. This approach eliminates the need for agents or complex inference pipelines, and allows for the generation of new scenarios using a few input images.

Recent 3D object generation methods achieve impressive high quality results by utilizing the sparse voxel representation~\cite{xiang2025structured, wu2025direct3d, he2025sparseflex, li2025sparc3d, lai2025lattice, xiang2025native}. However, these methods are constrained by the overall voxel grid resolution. In large scenes, the geometry becomes increasingly blurry and loses fine details due to the upper bound on the total number of voxels. Which is further exacerbated by uniformly distributed voxel grids, which allocate the same spatial resolution to empty and occupied regions alike limiting their use to large-scale scene generation as seen in~\cref{fig:trellis2_limitation}. In contrast, we represent the scene as a variable-length sequence of scene chunks. Each scene chunk is encoded as a flattened vector set \cite{zhang20233dshape2vecset}, as illustrated in \cref{fig:teaser}, resulting in a compact token sequence whose length corresponds to the number of scene chunks. Importantly, this flattened representation shifts part of the computation from the token length, which has quadratic memory growth, to the model width, which scales linearly. Because all the vector sets have a homogeneous size, they can be directly reshaped into token channels, enabling efficient scaling across scenes of different sizes while maintaining consistent fidelity.

To circumvent the resolution limit of object generators, another line of work~\cite{engstler2025syncity, zheng2025constructing3dtownsingle, chen2025trellisworld, yoon2025extendd} decomposes large scenes into chunks and leverages Trellis~\cite{xiang2025structured} to generate each chunk in a training-free manner. These approaches require multiple runs of Trellis sequentially or in parallel, leading to slow inference speeds and/or heavy memory usage. Moreover, some methods~\cite{engstler2025syncity} rely on additional resource-intensive modules, further increasing computational overhead. By comparison, our method is trained end-to-end and diffuses the entire scene at once during inference. This results in a significantly simpler inference pipeline and more efficient use of computational resources.

To summarize, our contributions are as follows: \textbf{Generative Boostrapping Strategy.} Our data generation pipeline combines a 3D reconstruction model with an expandable scene generation model, enabling the creation of scenes with various sizes and layouts from only a few input images of the target scenario. Paired with pseudo-sketches of these scenes, this pipeline supports end-to-end training of a controllable world model alleviating data scarcity. \textbf{Scalable Scene Representation.} We represent scenes as a variable-length sequence of scene chunk tokens, where each chunk is encoded as a flattened vector set. This yields a compact and efficient representation that scales with scene size while preserving consistent geometric fidelity. \textbf{Efficient End-to-End System.} Our model is trained end-to-end to generate complete scenes at once, eliminating the need for complex training-free pipelines or agent-based systems that may incur additional computational overhead.

\section{Related Work}

\subsection{Unbounded 3D Scene Generation} 
We review methods that perform expandable generation of arbitrarily large 3D scenes directly in the 3D domain. PDD~\cite{liu2024pyramid} uses a multi-scale pyramid diffusion framework to generate urban scenes. While several works~\cite{lee2024semcity, wu2024blockfusion, meng2025lt3sd} follow the latent diffusion paradigm, learning to generate chunks through diffusion in the latent space. Expandable generation is enabled with RePaint~\cite{lugmayr2022repaint} by conditioning on overlapping regions, requiring additional diffusion steps. SemCity~\cite{lee2024semcity} and BlockFusion~\cite{wu2024blockfusion} use triplane representations for urban and indoor scenes, respectively, while LT3SD~\cite{meng2025lt3sd} uses hierarchical feature grids for modeling finer indoor details. NuiScene~\cite{Lee_2025_ICCV} introduces a vector set representation for outdoor scenes and an explicit outpainting model for fast generation. WorldGrow~\cite{li2025worldgrow} fine-tunes Trellis~\cite{xiang2025structured} for 3D block inpainting, enabling block-by-block synthesis of scenes. The autoregressive nature of these methods causes inference time to scale with scene size. As shown in~\cref{tab:resource_compare}, NuiScene incurs higher runtime than models like ours that diffuses entire scenes at the same time.

\subsection{Training-Free Scene Generation}
\begin{figure*}[p]
\centering
\includegraphics[width=0.8\linewidth]{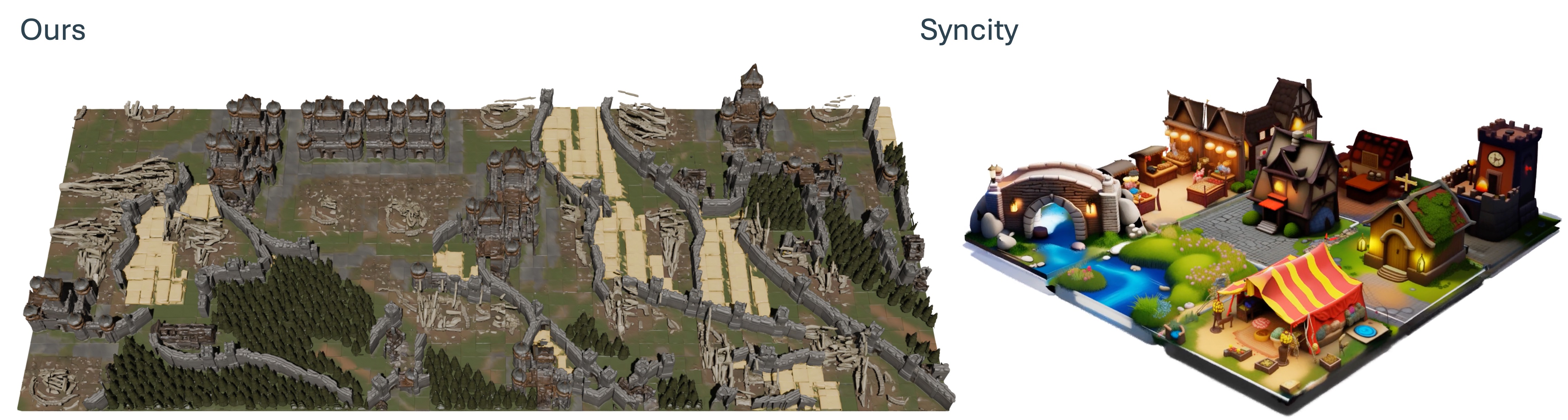}
\caption{Compared to training-free methods like SynCity~\cite{engstler2025syncity}, our scene (left) only took 5 minutes and 26 GB of VRAM to generate while SynCity took 54 minutes and 45 GB of VRAM.}
\label{fig:syncity_compare}
\end{figure*}

Several recent works leverage Trellis~\cite{xiang2025structured} to enable training-free, chunk-based generation, circumventing the resolution limits of object-centric generators. SynCity~\cite{engstler2025syncity} uses FLUX~\cite{flux2024} to generate smaller image tiles sequentially and Trellis to lift them into 3D. 3DTown~\cite{zheng2025constructing3dtownsingle} derives a point cloud from a reference image of the scene, and conditions Trellis on cropped images with corresponding cropped point clouds to generate local regions. TrellisWorld~\cite{chen2025trellisworld} and Extend3D~\cite{yoon2025extendd} introduces denoising schemes to enable parallel chunk generation for scenes. The primary limitation of these methods is their resource consumption, which is fundamentally bounded by the cost of running Trellis (\cref{tab:resource_compare}). Sequential chunk generation leads to inference time that scales linearly with scene size, whereas parallel generation comes at the expense of increasing memory usage. Our model is trained to generate whole scenes natively with latents that naturally scale with scene size (\cref{fig:syncity_compare}).

\subsection{LLM Aided Scene Generation} Recent works~\cite{lu2025yo, wang2025raisecity, huang2025majutsucity} leverage LLMs to assist city generation. Yo'City~\cite{lu2025yo} relies on LLMs for generating fine-grained text prompts for scene grids to drive image and subsequent 3D generation. RaiseCity~\cite{wang2025raisecity} derives buildings from street-view images via segmentation and used as inputs to image generation tools for completion before 3D generation and scene placement. MajutsuCity~\cite{huang2025majutsucity} learns to generate semantic layout and height maps from LLM enriched user input. Buildings are extruded with height maps, and used to condition image generation tools for image to 3D conversion and scene assembly. Across all methods, agents are further employed to iteratively evaluate and regenerate images of city regions or buildings before 3D generation. The inclusion of agent heavy pipelines can introduce longer generation times and API costs during inference. Additionally, these works mainly focus on cityscapes, it is unknown whether they can generalize to more open-domain scenarios. WorldGen~\cite{wang2025worldgen} relies on LLMs to produce initial procedural generation parameters from text input. The procedurally generated blockout is used as a condition for image generation. Followed by scene reconstruction, decomposition, and enhancement. Our sketch-to-world model operates end-to-end without LLMs and enables open-domain generation through our generative bootstrapping process.

\section{Preliminary}

\subsection{NuiScene}

In NuiScene~\cite{Lee_2025_ICCV}, the model is trained on $K$ scenes $\{\mathbf{S}_i\}_{i=1}^K$ processed from Objaverse~\cite{deitke2023objaverse}. Each scene is represented as $\mathbf{S}_i=(\mathbf{O}_i, \mathbf{P}_i)$, where $\mathbf{O}_i$ denotes an occupancy grid with dimensions $X_i \times Y_i \times Z_i$, and $\mathbf{P}_i\in\mathbb{R}^{N_{pc}\times3}$ is a point cloud with $N_{pc}$ points sampled from the marching cubes surface of $\mathbf{O}_i$.

Each scene is partitioned along the $x$ and $z$ axes into smaller chunks of fixed size $s\times Y_i \times s$, where $s < \min_i X_i$ and $s < \min_i Z_i$. We denote the chunk at location (u, v) as $\mathbf{S}_i^{(u,v)} = (\mathbf{O}_i^{(u,v)}, \mathbf{P}_i^{(u,v)})$, where $\mathbf{O}_i^{(u,v)}$ is the cropped occupancy and $\mathbf{P}_i^{(u,v)}\in\mathbb{R}^{N_{pc}^{(u,v)}\times3}$ is the corresponding point cloud from its surface. The model consists of a chunk VAE that encodes $\mathbf{P}_i^{(u,v)}$ into a vector set, and a diffusion model that learns over a local $2\times2$ grid of adjacent chunks.

\subsubsection{Chunk VAE} 
\citet{Lee_2025_ICCV} follows 3DShape2VecSet~\cite{zhang20233dshape2vecset} by employing a cross-attention (CA) layer to encode a scene chunk point cloud $\mathbf{P}_i^{(u,v)}$ into a vector set $\mathbf{z}\in\mathbb{R}^{V\times c}$, where $V$ is the number of vectors and $c$ is the channel size. The decoder then applies a stack of self-attention (SA) layers to get an output feature $\mathbf{f}_{out}\in\mathbb{R}^{F\times h}$. To increase the number of tokens, a fully connected (FC) layer may be added within the decoder paired with a pixel shuffle~\cite{shi2016real}-like up-sampling operation, producing $F$ tokens. Here $h$ denotes the model's hidden dimension. Finally, coordinates are sampled within each chunk's occupancy grid $\mathbf{O}_i^{(u,v)}$ and queried through an additional CA layer, followed by an FC layer, to produce occupancy logits. These predictions are supervised with the corresponding ground truth occupancy values $\mathbf{O}_i^{(u,v)}$.

\subsubsection{Quad Chunk Diffusion} The quad chunk diffusion model is trained to capture local $2\times2$ context windows of adjacent chunk latents, specifically $\mathbf{z}_i^{(u,v)}$, $\mathbf{z}_i^{(u,v+s)}$, $\mathbf{z}_i^{(u+s,v)}$, and $\mathbf{z}_i^{(u+s,v+s)}$. Training is performed using DDPM~\cite{ho2020denoising} with four different masking and conditioning configurations over this context window, enabling raster-scan order generation during inference for scenes of varying sizes and layouts. Please see the original paper for further details.

\section{Method}

In this section, we describe the NuiWorld framework. We begin by introducing our data generation pipeline in~\cref{sec:method-data_gen}. Obtaining initial 3D scenes to bootstrap the process (\cref{sec:method-bootstrap_3D}) to train NuiScene (\cref{sec:method-nuiscene}) and constructing a dataset of various scene sizes and layouts (\cref{sec:method-synth_3D}). Using this synthesized data, we then train our sketch-to-world model, as detailed in \cref{sec:method-sketch_to_world}.

\subsection{Data Generation}
\label{sec:method-data_gen}

\subsubsection{Bootstrapping Initial 3D Scenes}
\label{sec:method-bootstrap_3D}
We begin by bootstrapping our pipeline with image generation for a target scenario (e.g. medieval, desert, cyberpunk) using Nano Banana from Gemini 3~\cite{comanici2025gemini, gemini3_2025, nanobanana2025}. Specifically, we prompt the model to generate images of isometric scene chunks, as illustrated in~\cref{fig:pipeline}, with varying element compositions for each image. We then employ Trellis 2~\cite{xiang2025native} to reconstruct colored meshes from the generated images. We observe that images containing excessively large scenes lead to degraded geometry detail, therefore we restrict the amount of elements in the scene prompts. Finally, following NuiScene~\cite{Lee_2025_ICCV} we unify scene scales and ground thickness, converting the reconstructed scenes into point cloud and occupancy grids. This yields a bootstrapped dataset of $M$ scenes, $\mathbf{S}^{boot} = \{\mathbf{S}_i^{boot}\}_{i=1}^M$, $\mathbf{S}_i^{boot}=(\mathbf{O}^{boot}_i,\mathbf{P}^{boot}_i)$ which is used to train NuiScene. Note that prompt tuning with Gemini and scene preprocessing are time-consuming, and thus used only for bootstrapping. Please see the supplementary for additional visualizations.

\subsubsection{NuiScene Training}
\label{sec:method-nuiscene}
We follow the NuiScene training procedure using the bootstrapped scenes $\mathbf{S}^{boot}$. One key modification is that we retain the color information produced by Trellis 2, storing it in the point clouds $\mathbf{P}^{boot}_i\in\mathbb{R}^{N_{pc}\times6}$ which contains both the 3D coordinates and RGB values. During the training of the Chunk VAE we add an additional color prediction head implemented via CA and FC layers similar to the occupancy head. Color query coordinates and ground-truth color supervision are sampled from the corresponding chunk $\mathbf{P}^{boot}_{i,(u,v)}$, with RGB values supervised by an $\ell_{2}$ loss. During inference, we extract the mesh using marching cubes, sample points on the surface to query color predictions, and assign each mesh vertex the color of its nearest predicted point. Finally, we also replace DDPM~\cite{ho2020denoising} with rectified flow~\cite{lipman2022flow} for training the quad chunk diffusion model.

\subsubsection{Scene Synthesis Across Scales and Layout}
\label{sec:method-synth_3D}
Next we sample a dataset of $Q$ scenes $\mathbf{S}^{Nui}=\{\mathbf{S}^{Nui}_i\}_{i=1}^Q$ using the trained NuiScene model. For each scene, we first sample a target area $A\in[A_{min}, A_{max}]$ using log-uniform distribution. With probability $p_{square}=0.3$, we generate a square scene. Otherwise, we sample an aspect ratio $r \in [1.0, 3.0]$ from a log-uniform distribution to determine the scene dimensions $(R_i, C_i)$ accordingly, enforcing $C_i\geq R_i \geq15$. Here $R_i$ denotes the number of scene chunk rows, and $C_i$ corresponds to the number of chunks per row. We then render the extracted scene mesh from a fixed viewpoint at a resolution of $512\times512$ ensuring that scenes are elongated along the horizontal axis of the image, producing a colored and colorless rendering. Each rendering is converted using two methods: canny edge~\cite{canny2009computational} and \citet{chan2022learning}. This results in 4 sketches per scene. Each resulting scene consists of a 2D grid of vector sets and pseudo sketches denoted as $\mathbf{S}^{Nui}_i=(\mathcal{V}_i, \{\mathcal{I}^{(j)}_i\}_{j=1}^{4})$, where $\mathcal{V}_i,\in\mathbb{R}^{R_i\times C_i\times V\times c}$ and $\mathcal{I}^{(j)}_i\in\mathbb{R}^{512\times512\times1}$.

\subsection{Sketch to World Model}
\label{sec:method-sketch_to_world}

\subsubsection{Model Architecture}
Our model aims to generate scenes $\mathcal{V}$ with variable token length $R\times C$ and a channel size of $V\times c$, conditioned on a pseudo sketch input. Here, we drop the subscript $i$ for simplicity and consider a single training sample. During training we randomly sample a sketch $\mathcal{I}$ from $\{\mathcal{I}^{(j)}\}_{j=1}^{4}$. Note that we do not choose to concatenate vector sets into the token length like in AutoPartGen~\cite{chen2025autopartgen}, this is to avoid quadratic growth which would increase compute by $V^2$.

Our transformer model $\textbf{w}_{\phi}$ consists of $L$ blocks each containing a self-attention, cross-attention and feed forward network as illustrated in~\cref{fig:pipeline}. The sketch conditioning is incorporated through each cross-attention layer. Specifically the sketch $\mathcal{I}$ is encoded using a frozen DINOv2 encoder~\cite{oquab2023dinov2} to obtain the token embeddings  $\mathbf{z}_{sk}\in\mathbb{R}^{1374\times1024}$ and fed to cross-attention layers. 

We train our model using rectified flow~\cite{lipman2022flow}. During training we sample a timestep $t\in[0,1]$ and Gaussian noise $\bm{\epsilon}\sim\mathcal{N}(0,\mathbf{I})$, and construct noisy latents as $\mathcal{V}_{t} = (1-t)\mathcal{V}_{0} + t\epsilon$, where $\mathcal{V}=\mathcal{V}_0$ denotes the ground-truth scene latent.

Before feeding into the model, the noisy latents $\mathcal{V}_{t}$ are also added with positional embeddings and size embeddings. Each token is assigned a 2D spatial coordinate $(row, col)$, where $row\in[0, R)$ and $col\in[0, C)$ determined by the token's location in the scene, which are encoded with sinusoidal functions. In addition, a size embedding derived from $R\times C$ and also computed with sinusoidal encoding are shared across all tokens in the scene. Both embeddings are added to $\mathcal{V}_{t}$ as input to the transformer. 

The model $\mathbf{w}_{\phi}$ is trained with the objective:
\begin{equation}
    \mathbb{E}_{\mathcal{V}, z_{sk}, \bm{\epsilon}\sim\mathcal{N}(0, \mathbf{I}), t}\left[\|\mathbf{w}_{\phi}(\mathcal{V}_t,z_{sk},t)-(\epsilon - \mathcal{V}_0)\|^2_2\right]
\end{equation}
To enable classifier-free guidance during inference, the sketch conditioning $z_{sk}$ is randomly dropped with $0.2$ probability during training. The timestep $t$ is incorporated into the network via modulation, following Trellis~\cite{xiang2025structured}.

\subsubsection{Size Prediction} During inference, the scene dimensions $R$ and $C$ may not be available for a given input sketch. To address this, we learn an additional size prediction network conditioned on the input sketch. The network takes as input the CLS embedding $z_{cls}\in\mathbb{R}^{1024}$, extracted from DINOv2 given the sketch image $\mathcal{I}$, and predicts the scene dimensions $(\hat{R}, \hat{C})$. We supervise training with the ground truth layout $(R, C)$. As shown~\cref{fig:pipeline}, the predicted layout $(\hat{R}, \hat{C})$ can be used to initialize the number of noisy tokens for scene generation during inference. We note that the size prediction is optional and can also be specified by the user.
\section{Experiment}

For all experiments, we train a separate model for each scenario. This is mainly due to resource constraints and enables faster development. Training models on different scenarios independently allows us to iterate more quickly and avoid long training times.

\subsection{Dataset}

We construct datasets for three scenarios: medieval, desert, and cyberpunk. Each dataset is bootstrapped using images generated by Nano Banana and reconstructed with Trellis 2, producing $M=16, 12$, and $12$ initial bootstrapping scenes for each scenario, respectively. Please see the supplementary for more details. Using the boostrapped scenes, we train NuiScene with a chunk size of $s=60$.

For overfitting experiments, we generate $Q=720$ scenes using NuiScene for the medieval scenario, with scene area $A$ sampled from the range $[225, 625]$. For full set training, we generate $Q=8000$ scenes per scenario, with scene areas sampled from $[225, 1024]$. The datasets are split into $7200$ scene sketch pairs for training and $800$ scenes for validation. Each scene is generated with random seeds using NuiScene. For the overfitting experiments, metrics are calculated on the train set directly; for the full set experiments, metrics are calculated on the validation split. For even larger scenes, we fine-tune full set models on $1800$ additional scenes sampled from areas $[1024, 1600]$ and refer to these models as XL models in experiments. 

\subsection{Evaluation Metrics}

\subsubsection{NuiScene} We largely follow NuiScene's evaluation protocol. For the VAE, we report Chamfer Distance (CD), F-Score, and IoU. Following NuiScene, we sample 50k points per scene chunk and use a distance threshold of the voxel length ($2/60$) when computing CD and F-Score. For IoU, we similarly sample 10k points uniformly from occupied and unoccupied regions, along with 10k point sampled near the surface. In addition, we add a root mean squared error (RMSE) metric to evaluate the color prediction. Specifically, we sample 10k colored points from the surface of the scene chunk for this evaluation. Finally, for the diffusion model, we evaluate using the Fréchet PointNet++~\cite{qi2017pointnet++} Distance (FPD) and Kernel PointNet++ Distance (KPD). For these metrics, $2048$ points are sampled per quad chunk. For all metrics, we use 10k quad chunks for evaluation. Please see~\citet{Lee_2025_ICCV} for more details.

\subsubsection{NuiWorld} For NuiWorld, we evaluate sketch to world adherence and overall quality using the RMSE between predicted scene embeddings and ground truth embeddings $\mathcal{V}$ generated by NuiScene during the data generation process. Similarly, CD is calculated between the decoded mesh and the corresponding ground truth meshes from NuiScene. Unless otherwise specified, we use the ground truth scene dimension layout $(R, C)$ during inference.

\subsection{Ablation}
\label{sec:exp-ablation}

We find it important to compress the vector set dimensions as much as possible to enable better sketch-to-world performance. Accordingly, we ablate vector set \textbf{compression ratio} and the sketch-to-world \textbf{model width}, analyzing their impact on overfitting.

\subsubsection{NuiScene VecSet Compression Rate} 
We experiment with two different compression ratios, resulting in vector set sizes of $(V,c) = (16, 64)$ and $(8, 64)$. For the $(8, 64)$ configuration, we add a vector set upsampling layer, following~\citet{Lee_2025_ICCV}, to upsample the number of tokens to $512$ for the last 6 self-attention layers of the VAE decoder. This helps to compensate for the $2\times$ reduction in the number of vectors. More details are described in the supplementary.

We report quantitative evaluations for the VAE and quad chunk diffusion models in~\cref{tab:nuiscene_vae}. All metrics are computed using models trained on the medieval scenario. For the VAE, we additionally evaluate reconstruction on the scene chunks from the desert scenario as the validation set, which is unseen during training.

As shown in the tables, the $(16, 64)$ and $(8, 64)$ configurations achieve \textbf{comparable performance}, with only negligible differences. This indicates that the additional compression incurs little to no loss of performance. We do note, however, that the $(8, 64)$ model is slightly slower and more memory-intensive to train due to the added upsampling layers, but is a worthwhile trade-off.

\begin{table}
\centering
\caption{Quantitative comparison of reconstruction and generation for quad-chunks across different VAE compression ratios. Here $\hat{h}$ indicates that the predicted height was used for occupancy prediction and $h$ the ground truth height as in NuiScene~\cite{Lee_2025_ICCV}. KPD scores are multiplied by $10^3$.}
\vspace{-1em}
\resizebox{\linewidth}{!}
{
\begin{tabular}{@{}lrrrrrr|rr@{}}
\toprule
 & \multicolumn{6}{c|}{\textbf{Reconstruction}} 
 & \multicolumn{2}{c}{\textbf{Generation}} \\
\cmidrule(lr){2-7} \cmidrule(lr){8-9}
VecSet Size 
& IOU$\uparrow$
& RGB RMSE$\downarrow$
& CD $(\hat{h})\downarrow$
& F-Score $(\hat{h})\uparrow$
& CD $(h)\downarrow$
& F-Score $(h)\uparrow$
& FPD$\downarrow$ 
& KPD$\downarrow$ \\
\midrule

\multicolumn{9}{c}{\textit{Medieval (Seen)}\xspace} \\
\addlinespace[0.3em]
(16, 64) & 0.950 & 0.028 & 0.048 & 0.828 & 0.048 & 0.827 & 1.430 & 2.893 \\
(8, 64) & \textbf{0.959} & \textbf{0.028} & \textbf{0.048} & \textbf{0.830} & \textbf{0.048} & \textbf{0.830} & \textbf{1.391} & \textbf{2.748}\\
\midrule
\multicolumn{9}{c}{\textit{Desert (Unseen)}\xspace} \\
\addlinespace[0.3em]
(16, 64) & 0.587 & 0.176 & \textbf{0.179} & 0.474 & \textbf{0.179} & \textbf{0.470} & -- & -- \\
(8, 64) & \textbf{0.589} & \textbf{0.172} & 0.184 & \textbf{0.474} & 0.185 & 0.467 & -- & -- \\
\bottomrule
\end{tabular}
}
\vspace{-1em}
\label{tab:nuiscene_vae}
\end{table}

\subsubsection{Sketch to World Model Width}
In~\cref{tab:overfit_width}, we show the effect of the sketch to world model width (hidden dimension) in relation to the vector set compression ratios. Specifically, when the model width is set equal to the flattened vector set dimension, i.e. $w=V \times c$ (rows 1 and 3 in the table), the model struggles to overfit to the $720$ scenes, as reflected by higher RMSE and CD. In contrast, increasing the model width to $w=1536 > V \times c$ (row 2 and 4) leads to a substantial performance improvement. The effect is particularly pronounced for the $(8, 64)$ configuration, where the gap between the model width and the flattened vector dimension is larger. This observation is in line with the recent work RAE~\cite{zheng2025diffusion}. Note that due to resource constraints we reduce the model depth to $16$ layers for $w=1536$, allowing training to fit on 2 L40S GPUs.

\begin{figure}
\centering
\includegraphics[width=\linewidth]{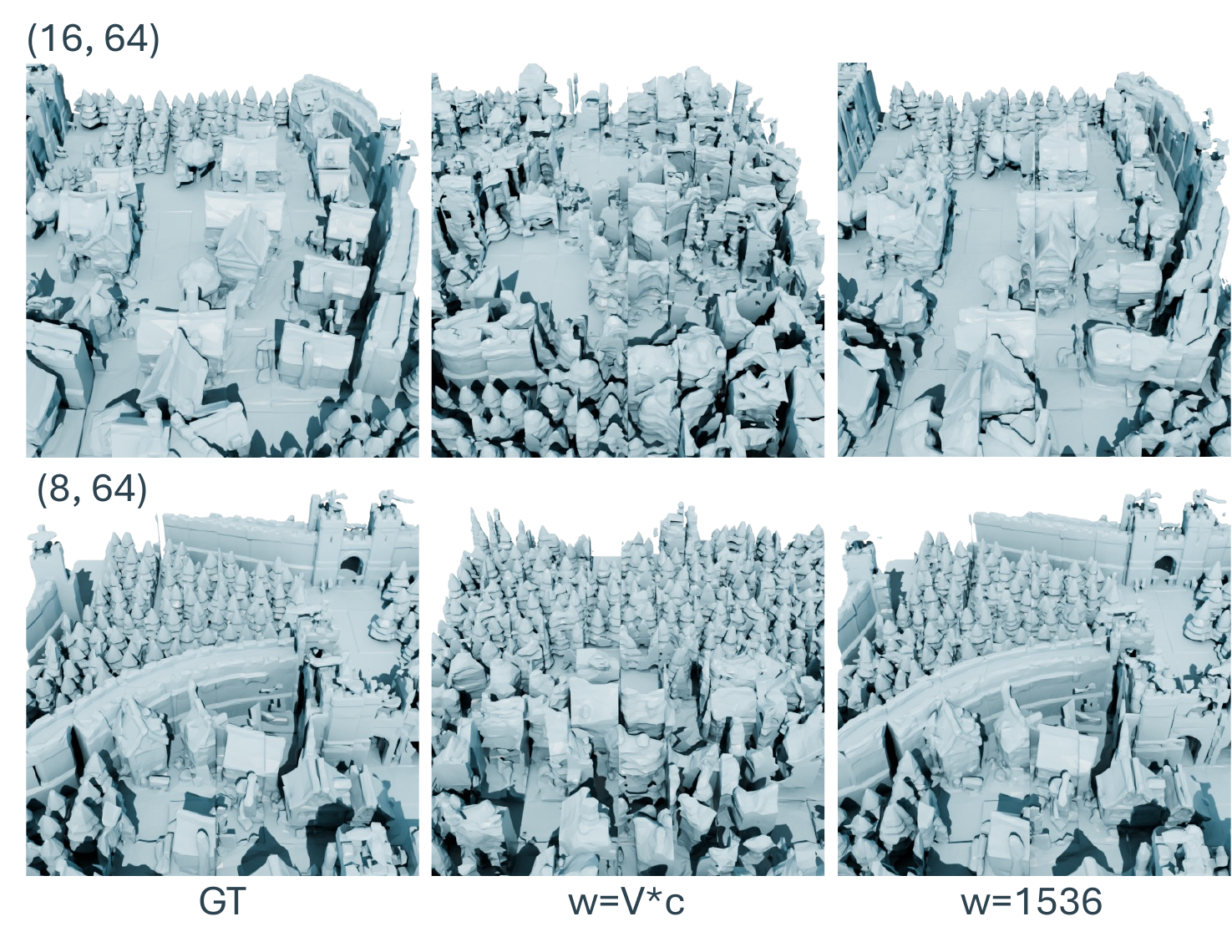}
\vspace{-2em}
\caption{Qualitative comparison of ground-truth NuiScene scenes and generated outputs from sketch-to-world models on the medieval overfit set across different VecSet compression ratios and model widths.}
\label{fig:qual_nuiworld_width}
\vspace{-5pt}
\end{figure}

We show qualitative results in~\cref{fig:qual_nuiworld_width}. When the model width is set to $w=V \times c$ (column 2), both configurations generate noisy chunks that are not coherently assembled. In contrast, with a wider model ($w=1536$, column 3), both models produced substantially more coherent scenes. The $(16, 64)$ model exhibits slightly degraded quality compared to the ground-truth scenes. While the $(8, 64)$ model more effectively overfits to the training data. This highlights the importance of having \textbf{higher compression ratios}, allowing for a larger margin between model width and flattened vector set dimension leading to \textbf{better performance} overall.

\begin{table}
\centering
\caption{Evaluation of the impact of compression ratio and model width on memory consumption and performance of the sketch-to-world model using the medieval overfit set.}
\vspace{-1em}
\resizebox{\linewidth}{!}
{
\begin{tabular}{@{}llrrrrr@{}}
\toprule
VecSet Size & BS & Width & Depth & VRAM & RMSE$\downarrow$ $(R, C)$ & CD$\downarrow$ $(R, C)$ \\
\midrule
(16, 64) & 24 & 1024 & 24 & 40 GB & 0.270 & 0.591 \\
(16, 64) & 24 & 1536 & 16 & 70 GB & 0.150 & 0.266 \\
(8, 64) & 24 & 512 & 24 & 18 GB & 0.312 & 0.663 \\
(8, 64) & 24 & 1536 & 16 & 70 GB & \textbf{0.078} & \textbf{0.161} \\
\bottomrule
\end{tabular}
}
\label{tab:overfit_width}
\vspace{-1em}
\end{table}

\begin{table}
\centering
\caption{Full-set evaluation on the medieval validation set across different scene size ranges, with CD also reported for scene layouts predicted by the scene size predictor.}
\vspace{-1em}
\resizebox{0.8\linewidth}{!}
{
\begin{tabular}{@{}lrrr@{}}
\toprule
Scene Size & RMSE$\downarrow$ $(R, C)$ & CD$\downarrow$ $(R, C)$ & CD$\downarrow$ $(\hat{R}, \hat{C})$ \\
\midrule
$[225, 625)$ & 0.210 & 0.337 & 0.904 \\
$[625, 1024]$ & 0.241 & 0.449 & 1.418 \\
$[225, 1024]$ & 0.220 & 0.373 & 1.070 \\
\bottomrule
\end{tabular}
}
\label{tab:fullset}
\vspace{-1em}
\end{table}

\begin{figure*}
    \centering
    \vspace{-1em}
    \includegraphics[width=\linewidth]{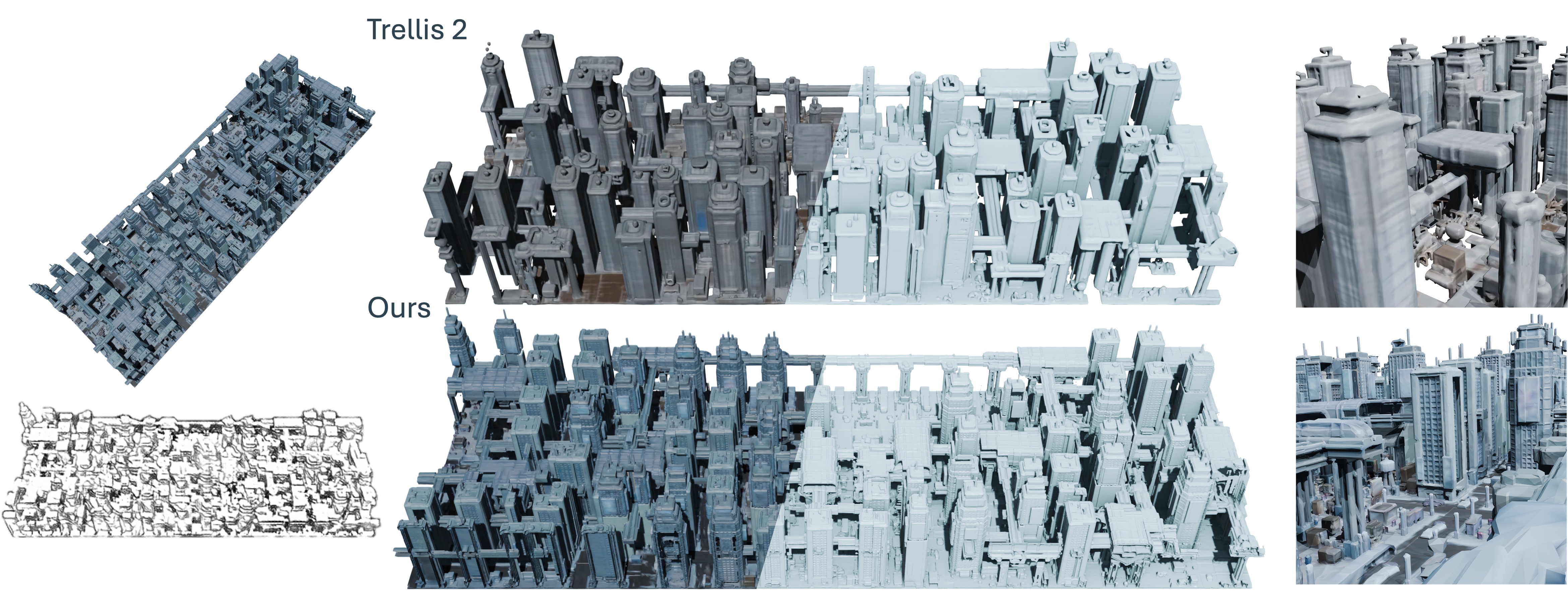}
    \caption{Qualitative comparison with Trellis 2. Trellis 2 takes as input a rendered view of the scene generated by our sketch-to-world model (top left), which itself is conditioned on the sketch shown in the bottom left. To better utilize the $1024\times1024$ input resolution for Trellis 2 and minimize detail loss for elongated scenes, we rotate the scene during rendering. The center column shows the fully generated scenes, while the right column presents zoomed-in renderings.}
    \label{fig:trellis2_compare}
    \vspace{-1em}
\end{figure*}

\subsection{Full Set Training}
\label{sec:exp-full_set}

We show full set medieval training results in~\cref{tab:fullset} using the best settings from the ablation, i.e., $(8, 64)$ VecSet size, model width$=1536$, depth$=16$, and batch size of $24$. We observe higher RMSE and CD scores for larger scene sizes (row 2 compared to row 1). This is because sketches have $512^2$ resolution with larger scenes having coarser sketches, so the model needs to infer plausible content in those regions that may not be 100\% aligned with the ground truth scene contributing to the the larger values.

When using predicted scene layouts $(\hat{R}, \hat{C})$ using the scene size predictor, we do notice a drop in performance. This is due to a relatively small training size of $7200$ and the low sketch resolution leading to inaccurate size predictions. However, we note that each scene chunk spans a length of $2$ in the global space, so the average CD still remains within the extent of a single scene chunk.

\subsubsection{Trellis 2 Comparison}

\cref{fig:trellis2_compare} shows results on the full set cyberpunk scenario compared with Trellis 2~\cite{xiang2025native}. We observe artifacts caused by their uniform voxel grid allocation (\cref{fig:trellis2_limitation}) results in holes in the generated meshes. In contrast, our model maintains \textbf{consistent details} even for larger scenes.

We show runtime statistics in~\cref{tab:resource_compare}. For Trellis~\cite{xiang2025structured} and Trellis 2, we benchmark two inputs: the cyberpunk scene generated by Nano Banana (\cref{fig:pipeline}) and the large scene in~\cref{fig:trellis2_compare}. For NuiScene and our method, we pick another sketch from the validation set with $15\times15$ size with the same scene elements. The scene sizes here are measured in our chunk sizes, as the output scenes do not have an absolute metric scale.

For Trellis and Trellis 2, the number of sparse voxels (tokens) during inference decreases for larger scenes (column 3), due to the aspect ratio and size. Resulting in the loss of detail and fidelity. In contrast, the \textbf{number of tokens in our model scales with the scene size}, preserving a consistent level of detail. One limitation is our decoding time (column 6), however, this may be improved with techniques like FlashVDM~\cite{lai2025unleashing}.

\begin{table}
\centering
\caption{Resource usage and generation statistics for different methods, decode time measures the time to convert the generated embeddings into mesh. Here, scene size refers to the spatial extent of the scene depicted in the input image or sketch, expressed in terms of the chunk dimensions.}
\vspace{-1em}
\resizebox{\linewidth}{!}
{
\begin{tabular}{@{}llrrrrrrr@{}}
\toprule
Scene Size & Method & \# Token & VRAM & Emb Gen Time (s) & Decode Time (s) \\
\midrule
$12\times13$ & Trellis & 20014 & 16 GB & 9.94 & 58.25 \\
$18\times51$ & Trellis & 8713 & 13 GB & 8.64 & 25.34 \\
$12\times13$ & Trellis 2 & 18340 & 34 GB & 55.19 & 84.94 \\
$18\times51$ & Trellis 2 & 8488 & 28 GB & 22.42 & 72.15 \\
\midrule
$15\times15$ & NuiScene & -- & 7 GB & 18.92 & 62.00 \\
$18\times51$ & NuiScene & -- & 7 GB & 47.19 & 278.76 \\
$15\times15$ & Ours & 225 & 19 GB & 5.96 & 64.61 \\
$18\times51$ & Ours & 918 & 19 GB & 5.56 & 276.76 \\
\bottomrule
\end{tabular}
}
\label{tab:resource_compare}
\vspace{-1em}
\end{table}

\begin{figure}
\centering
\includegraphics[width=\linewidth]{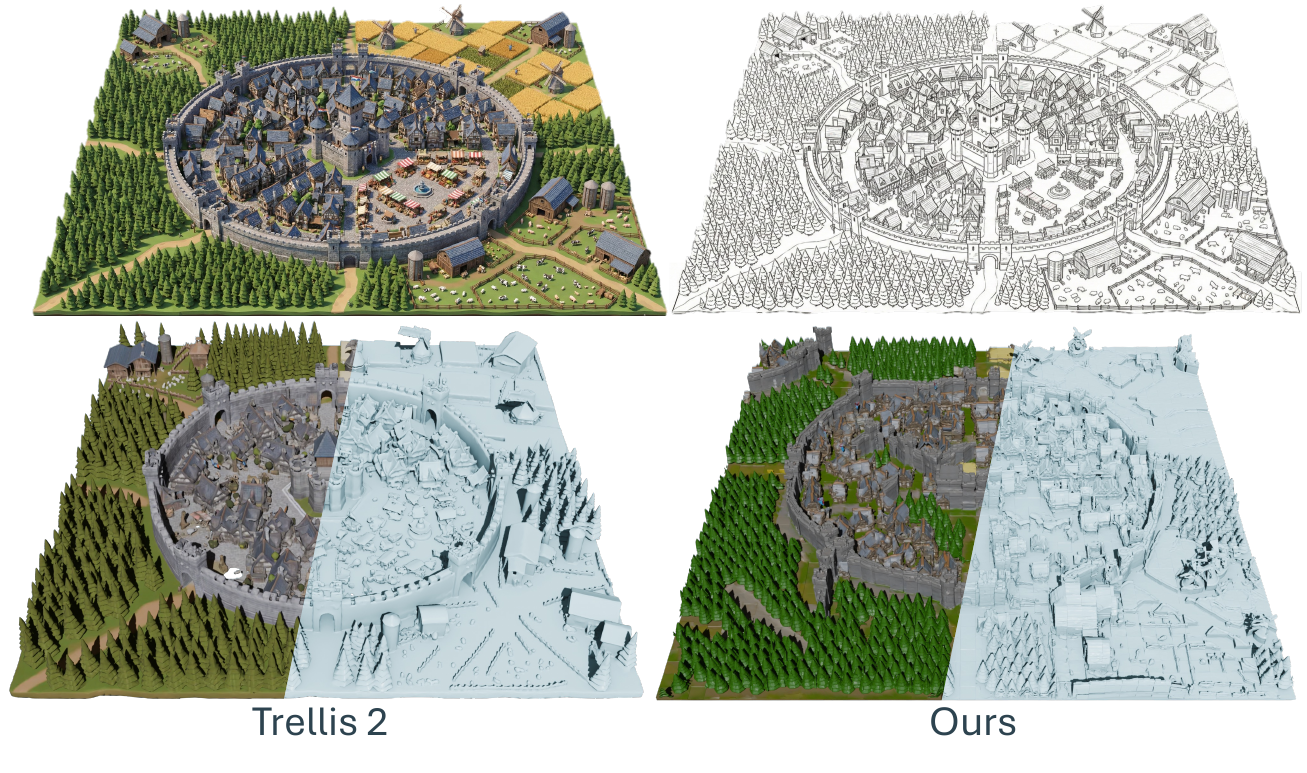}
\vspace{-20pt}
\caption{The top row shows a large medieval scene image and its corresponding sketch both generated by Nano Banana. The bottom row presents the results produced by Trellis 2 and our XL model using the above as input. Zoom in for more details.}
\label{fig:unseen_sketch}
\vspace{-2em}
\end{figure}

\subsection{Generalization to Unseen Sketch}

In \cref{fig:unseen_sketch}, we demonstrate our method’s generalization to unseen sketches by using Gemini to convert a large scene image into a sketch. For this experiment, we use the XL model trained on the medieval scenario, and we set $(R, C)=(40, 40)$ during inference. As shown, our model \textbf{exhibits a degree of generalization}, capturing structures such as circular walls and the corresponding wheat fields and farm barns in the relevant sketch regions. While the generated scene does not perfectly adhere to the input sketch, the result is notable given that our method is trained on a relatively small number of scenes and does not encounter sketches of this style or layout in the ground-truth dataset. Compared to Trellis 2, our model still lacks global scene coherence and exhibits seam artifacts and noisy regions, which are from the NuiScene-generated training data.

\section{Limitation}

\mypara{Scaling.} Our method's generalizability is currently limited by the modest training set size. Nevertheless, we view this work as a proof of concept for addressing the \textbf{scalability} and \textbf{data scarcity} challenges in world generation. The local detail quality currently also remains inferior. This is mainly due to limitations in our model capacity. However, we highlight this is largely due to resource constraints and may be addressed by scaling up model capacity and training data with additional computational resources.

\begin{figure*}[p]
\centering
\includegraphics[width=0.9\linewidth]{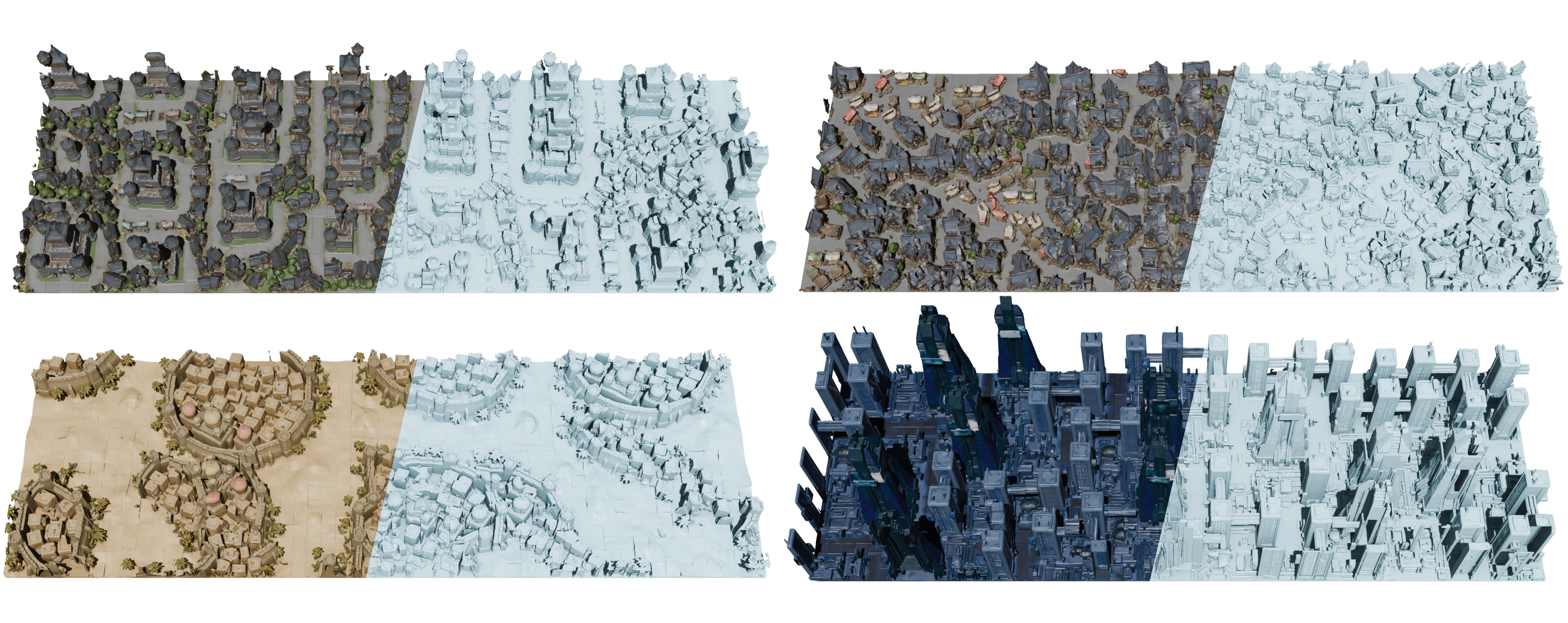}
\caption{Example scenes generated by NuiScene during the data generation pipeline for the medieval, desert and cyberpunk scenarios.}
\label{fig:failure}
\end{figure*}

\mypara{Layout.} Due to the resolution limits of Trellis 2, the scale of scenes that can be produced from bootstrapping is constrained. Combined with color shifts and geometric scale inconsistencies in the reconstructions, we find that NuiScene's generated scenes tend to draw most of the content from the same single bootstrapped training scene, as shown in~\cref{fig:failure}. We also observe that seams between chunks are quite visible in NuiScene generations. That being said, these components in our framework are modular and can be easily swapped out for better models in the future.

\mypara{Monolithic Scene.} For applications like simulation, decomposability of scene elements is an important requirement. Currently, our approach generates scenes in a monolithic fashion, producing a single fused mesh as the final output. WorldGen~\cite{wang2025worldgen} has demonstrated that it is possible to decompose monolithic scenes by leveraging recent part generation techniques~\cite{chen2025autopartgen, tang2025efficient}. We believe that applying similar decomposition methods to our generated outputs would be straightforward.

\section{Conclusion}
In this work, we present NuiWorld, a framework designed to address two fundamental challenges in world generation: \textbf{large scale} and \textbf{open-domain} scenario generation. To enable large scale scene synthesis, we introduce a variable-length scene chunk representation that adapts to scene size, preserving consistent fidelity. We also propose a flattened vector-set token representation that shifts computation from token length to the channel dimension, improving the training efficiency. For open-domain generation, we develop a pipeline that leverages off-the-shelf image, object and expandable scene generation methods to create scenes of varying scenarios, sizes and layout. This allows for end-to-end controllable generation with pseudo sketches. While our method is currently limited in scale, it provides clear advantages over prior methods and provides useful insights toward fast and scalable world generation.

\vspace{1em}

\mypara{Acknowledgements.}
This work was funded by a CIFAR AI Chair, an NSERC Discovery grant, and enabled by support from the \href{https://alliancecan.ca/}{Digital Research Alliance of Canada} and a CFI/BCKDF JELF grant. We thank Jiayi Liu for helpful suggestions on improving the paper.

\bibliographystyle{ACM-Reference-Format}
\bibliography{main}

\clearpage

\begin{figure*}[p]
\centering
\includegraphics[width=\linewidth]{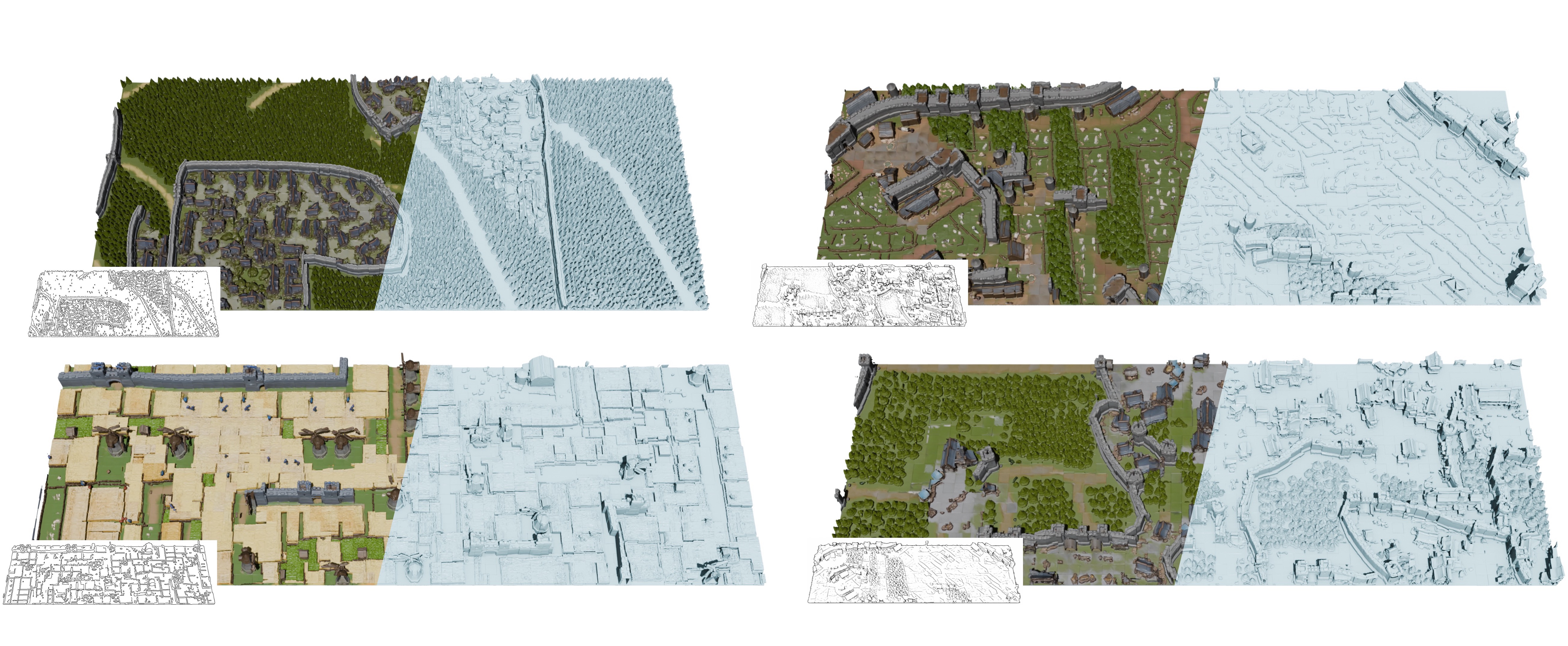}
\caption{Example scenes generated by our XL sketch to world model for the medieval scenario, input sketches are from the validation set.}
\label{fig:medieval_extra}
\end{figure*}

\begin{figure*}[p]
\centering
\includegraphics[width=\linewidth]{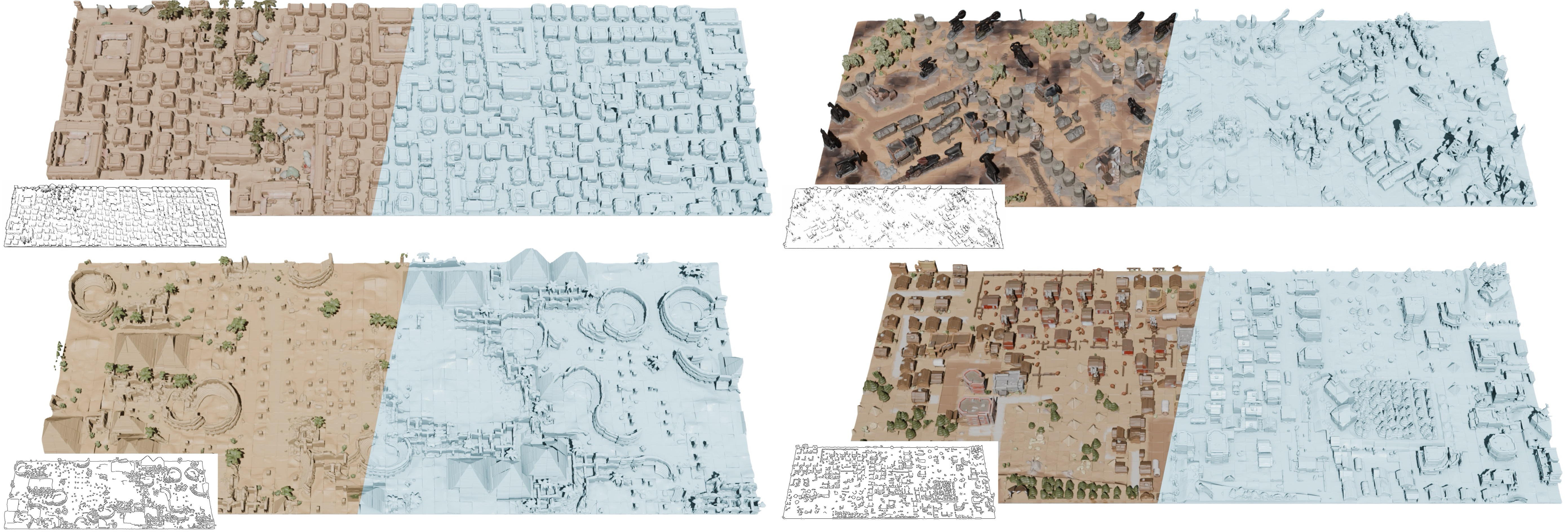}
\caption{Example scenes generated by our XL sketch to world for the desert scenario, input sketches are from the validation set.}
\label{fig:desert_extra}
\end{figure*}

\begin{figure*}[p]
\centering
\includegraphics[width=\linewidth]{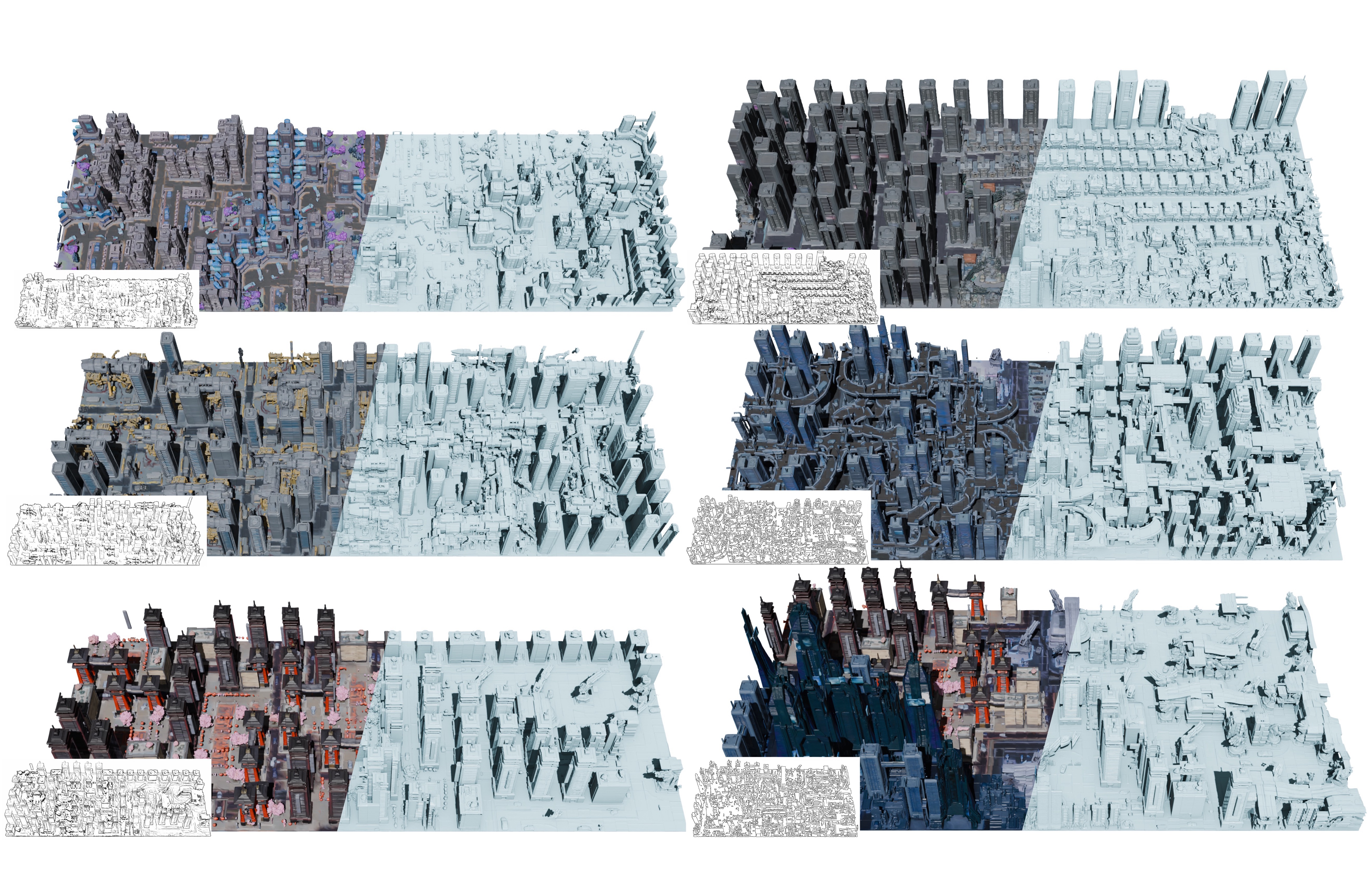}
\caption{Example scenes generated by our XL sketch to world model for the cyberpunk scenario, input sketches are from the validation set.}
\label{fig:cyberpunk_extra}
\end{figure*}

\clearpage

\appendix
\section{Dataset}
\label{sec:appendix_dataset}

\begin{figure*}
\centering
\vspace{-1em}
\includegraphics[width=\linewidth]{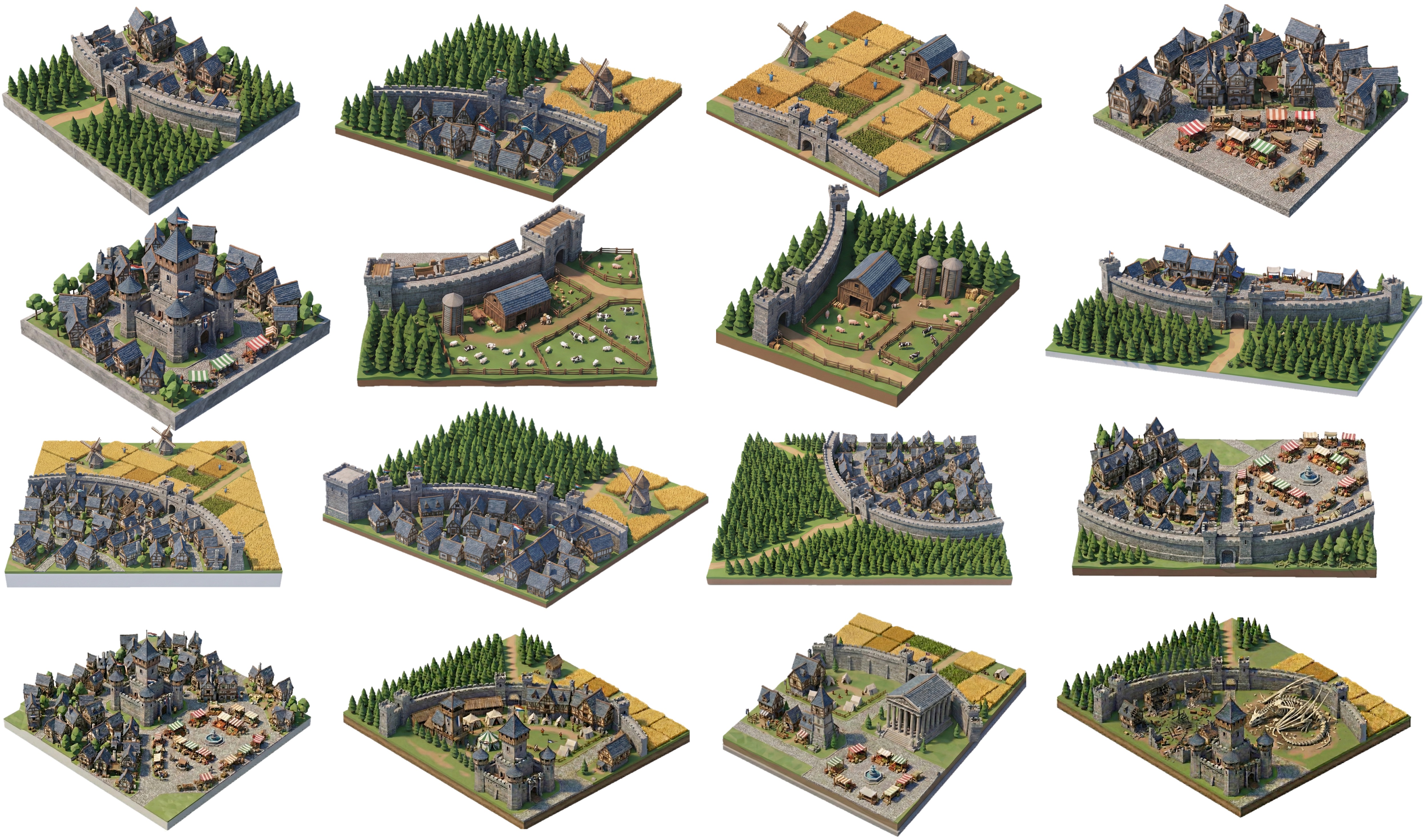}
\caption{Images generated for the medieval scenario using Nano Banana.}
\label{fig:medievalv2_gemini}
\end{figure*}

\begin{figure*}
\centering
\vspace{-1em}
\includegraphics[width=\linewidth]{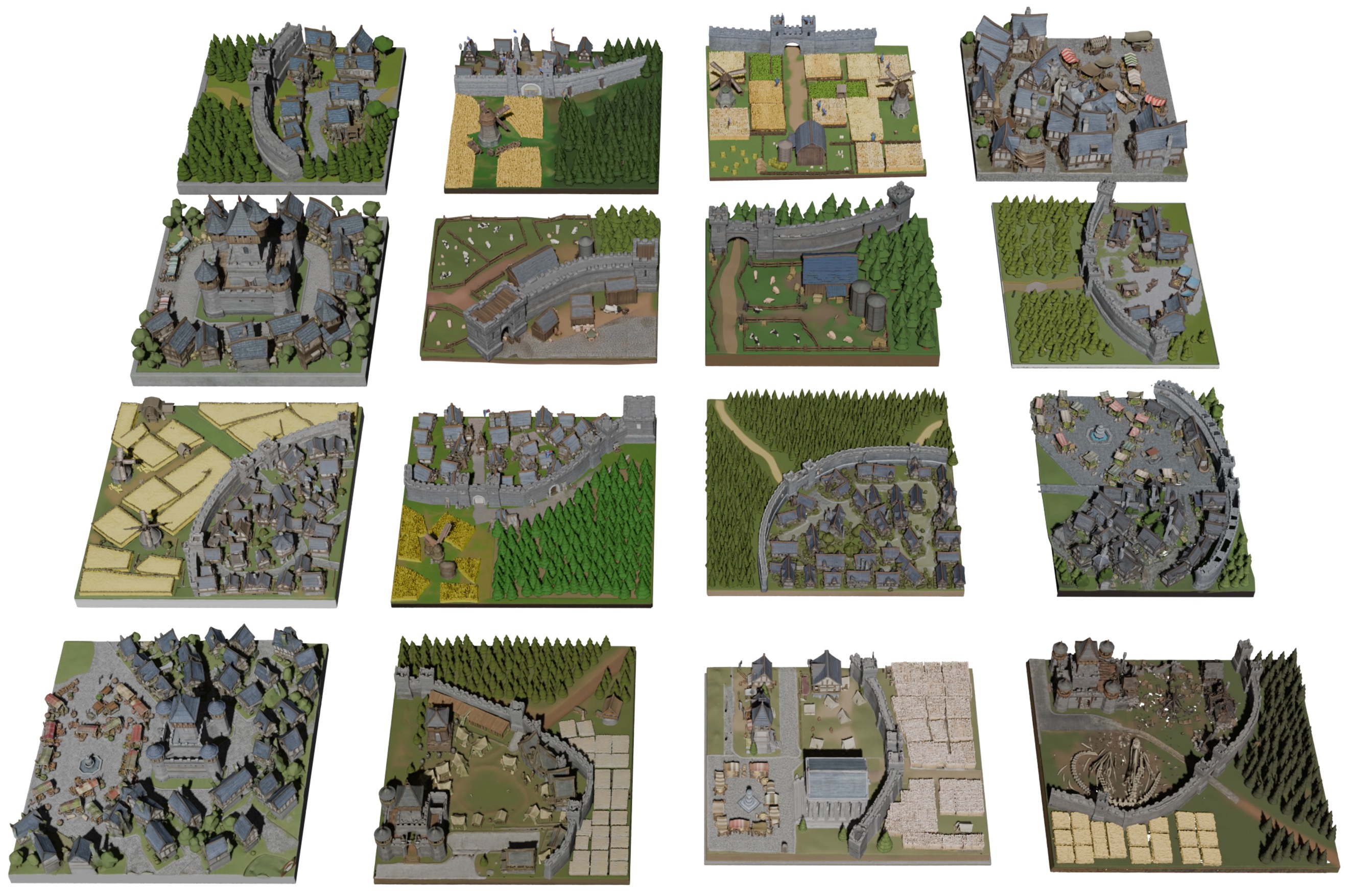}
\caption{Scenes reconstructed for the medieval scenario using Trellis 2.}
\label{fig:medievalv2_trellis2}
\end{figure*}

\begin{figure*}
\centering
\vspace{-1em}
\includegraphics[width=\linewidth]{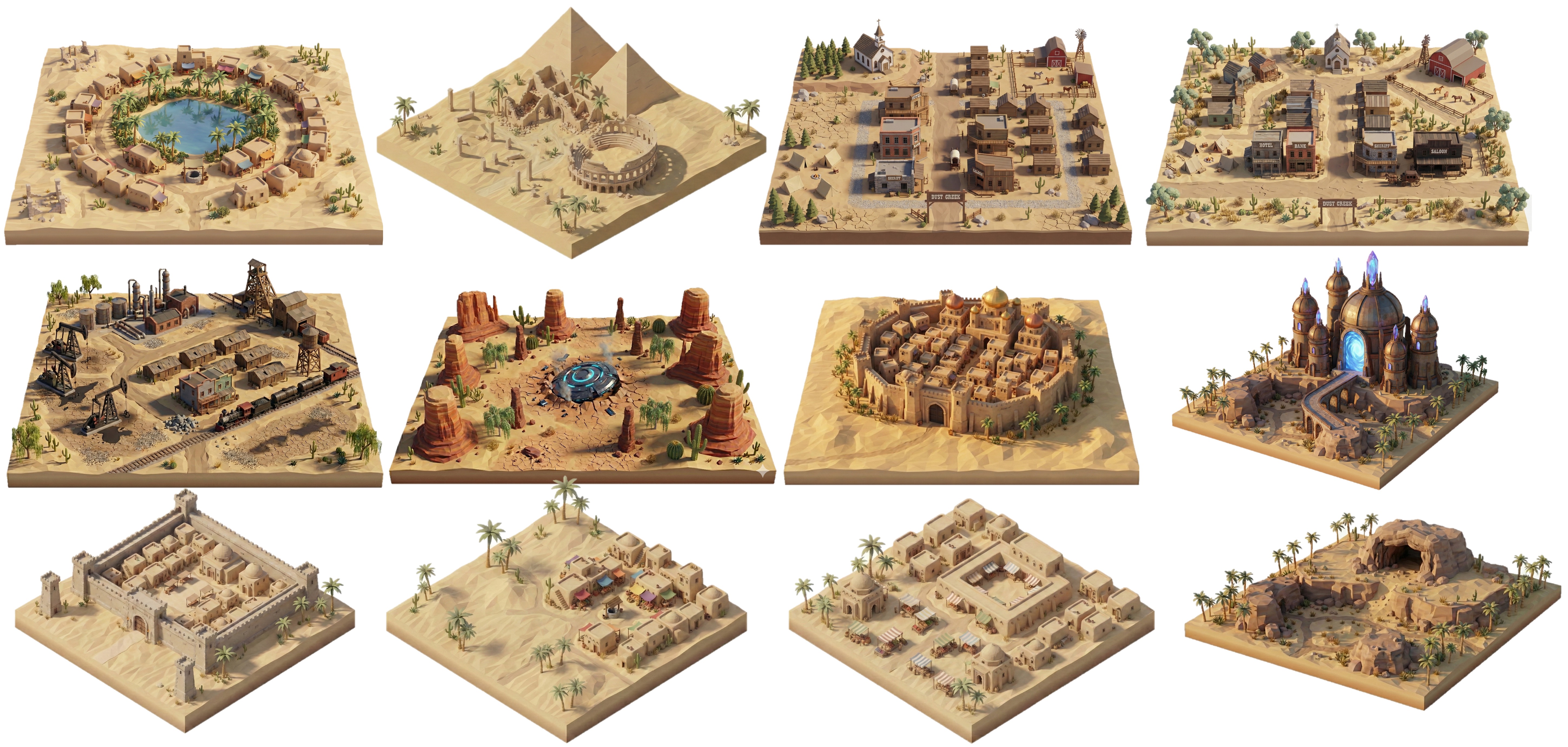}
\caption{Images generated for the desert scenario using Nano Banana.}
\label{fig:desertv2_gemini}
\end{figure*}

\begin{figure*}
\centering
\vspace{-1em}
\includegraphics[width=\linewidth]{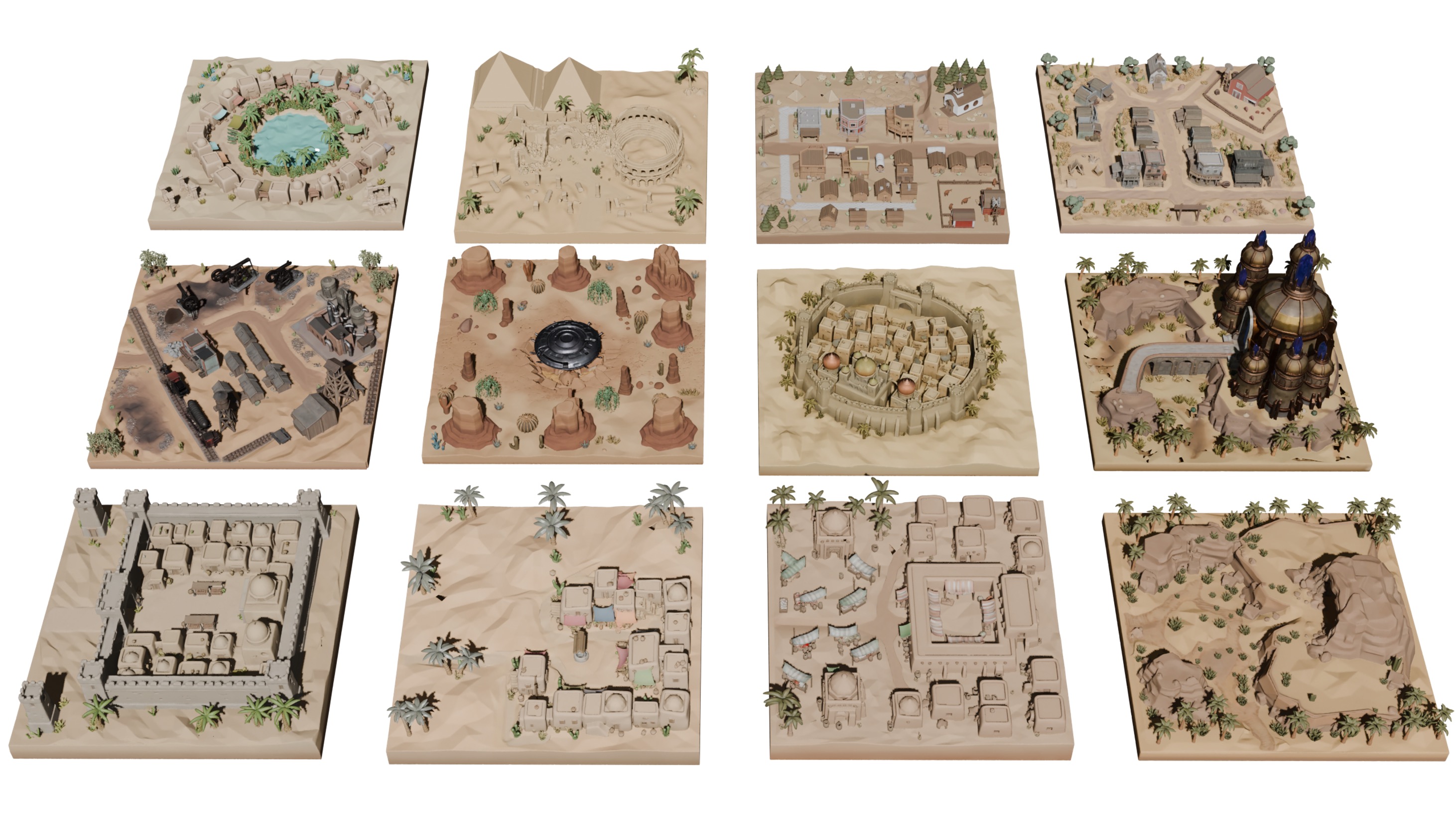}
\caption{Scenes reconstructed for the desert scenario using Trellis 2.}
\label{fig:desertv2_trellis2}
\end{figure*}

\begin{figure*}
\centering
\vspace{-1em}
\includegraphics[width=\linewidth]{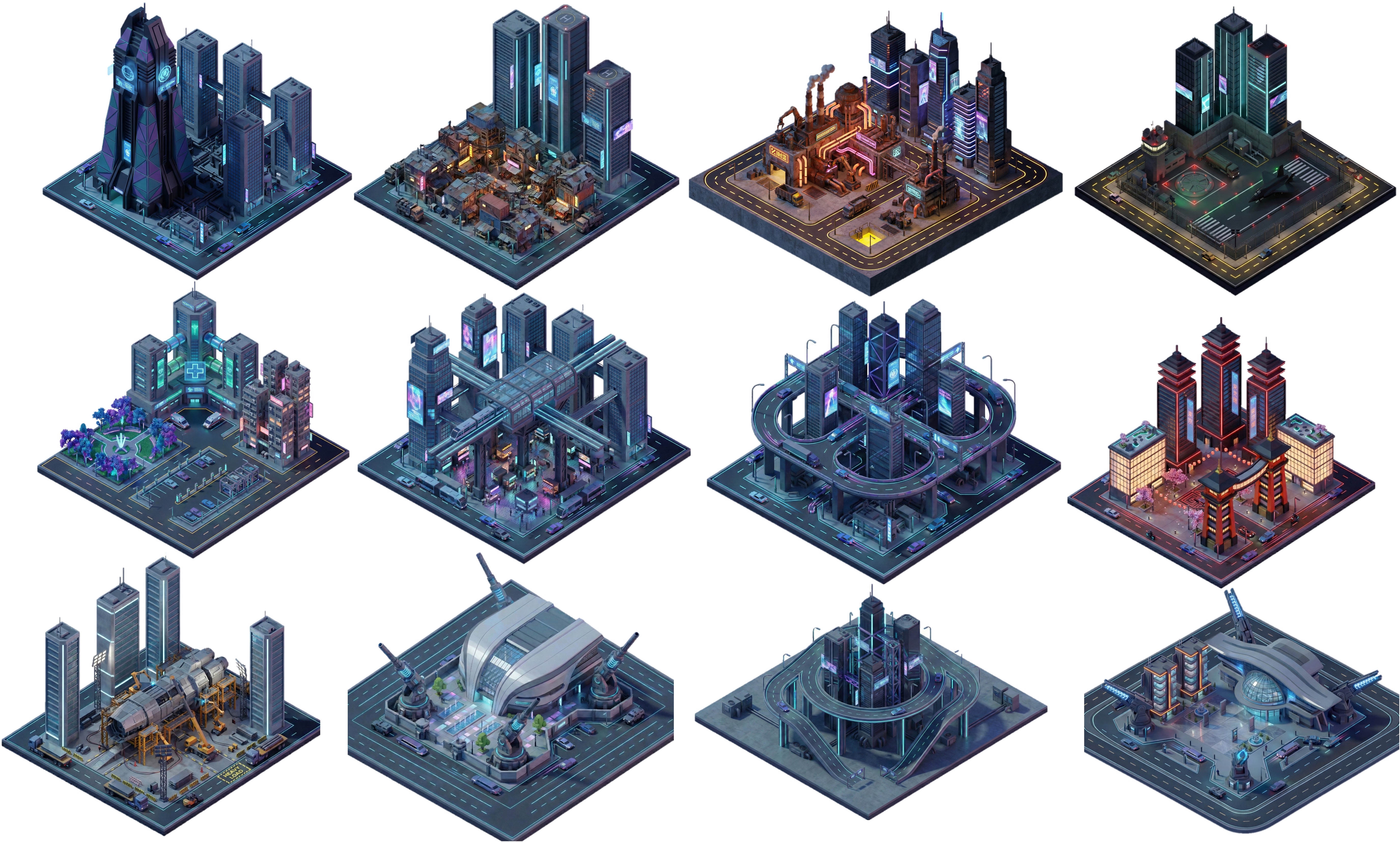}
\caption{Images generated for the cyberpunk scenario using Nano Banana.}
\label{fig:cyberpunk_gemini}
\end{figure*}

\begin{figure*}
\centering
\vspace{-1em}
\includegraphics[width=\linewidth]{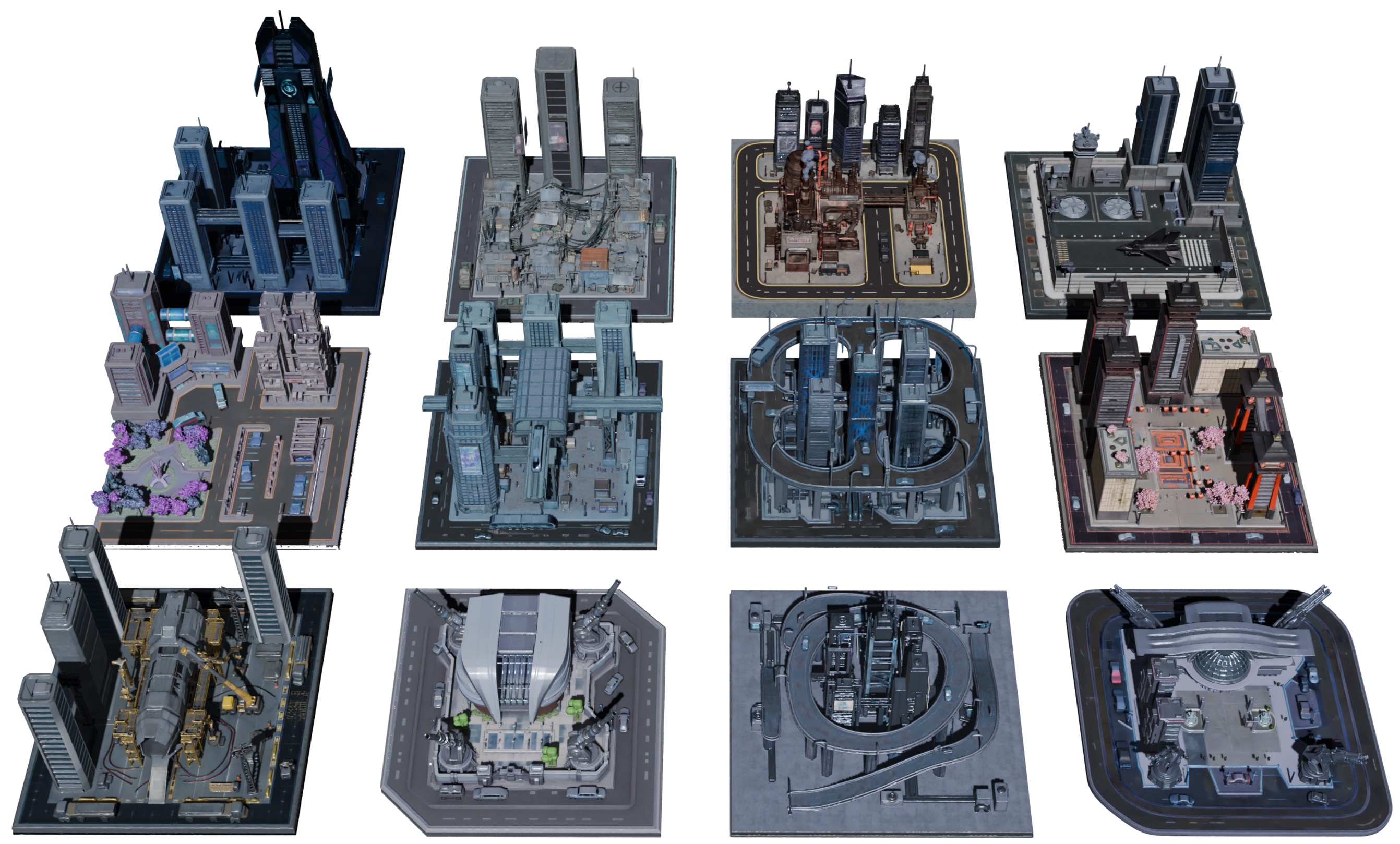}
\caption{Scenes reconstructed for the cyberpunk scenario using Trellis 2.}
\label{fig:cyberpunk_trellis2}
\end{figure*}

We show the images generated using Nano Banana~\cite{comanici2025gemini, gemini3_2025, nanobanana2025} in~\cref{fig:medievalv2_gemini,fig:desertv2_gemini,fig:cyberpunk_gemini}, and corresponding reconstructed scenes using Trellis 2~\cite{xiang2025native} in~\cref{fig:medievalv2_trellis2,fig:desertv2_trellis2,fig:cyberpunk_trellis2} for the medieval, desert and cyberpunk scenarios, respectively.

\subsubsection{Prompting Strategy} 
\begin{figure*}
\centering
\vspace{-1em}
\includegraphics[width=\linewidth]{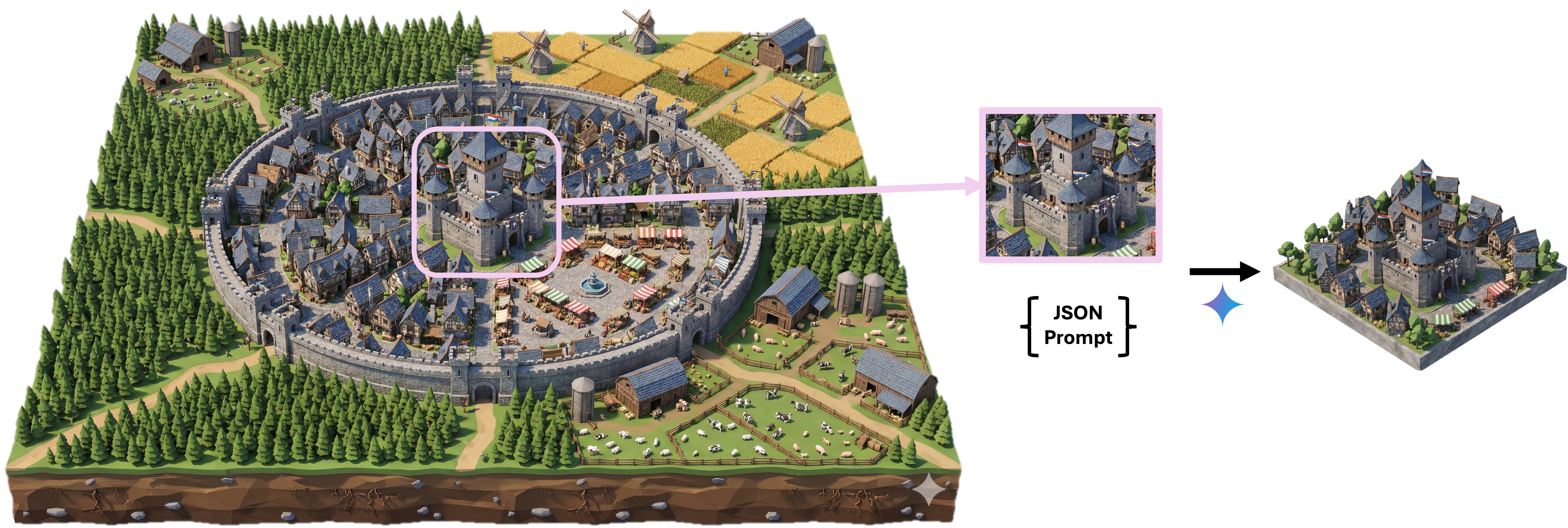}
\caption{Prompting strategy to takes a large scene image, extract a smaller crop, combined with a JSON instruction prompt to guide Nano Banana to generate a clean, isolated isometric chunk image.}
\label{fig:prompt_strategy1}
\end{figure*}

We employ two main strategies for prompting Nano Banana to generate isolated isometric scene chunks. The first strategy involves generating an image of a very large scene. Then we use crops of the scene to further prompt Nano Banana to create smaller scene chunks. An example illustration is shown in~\cref{fig:prompt_strategy1} and corresponding JSON prompt in~\cref{lst:prompt1}.

\begin{lstlisting}[language=json, caption={Prompt Strategy 1}, label={lst:prompt1}]
{
"task": "img2img_constrained_chunking",
"critical_instructions": {
"size_constraint": "NO EXPANSION. Do not generate new landscape beyond the input crop. The scene footprint must remain compact and limited to the input objects.",
"geometry_constraint": "RECTANGULAR BASE. The ground plinth must be a sharp, geometric rectangle. No circular or organic shapes."
},
"scene_metadata": {
"style": "Low-poly 3D render, miniature diorama",
"perspective": "45° top-down isometric, wide view",
"composition": "A compact, isolated scene chunk on a rectangular slab, fully visible against white.",
"background": "Solid white"
},
"generation_parameters": {
"positive_prompt": "A compact, self-contained isometric diorama chunk floating in a white void. The scene sits on a sharp, perfectly RECTANGULAR ground slab. The geometry is tightly fitted to the input content-repairing cut edges of walls and trees without adding extra landscape or sprawling terrain. Wide shot, fully visible with white padding on all sides. High definition low-poly 3D render.",
"negative_prompt": "expanding landscape, sprawling terrain, added buildings, circular base, round plinth, oval base, organic shape base, zoomed in, cut off, cropping, touching frame, missing architecture, messy edges",
"denoising_strength": 0.5,
"guidance_scale": 9.0
},
"notes": "The phrase 'tightly fitted' tells the model to wrap the base closely around the existing house/trees rather than creating a large island with empty space."
}
\end{lstlisting}

\begin{figure}
\centering
\includegraphics[width=\linewidth]{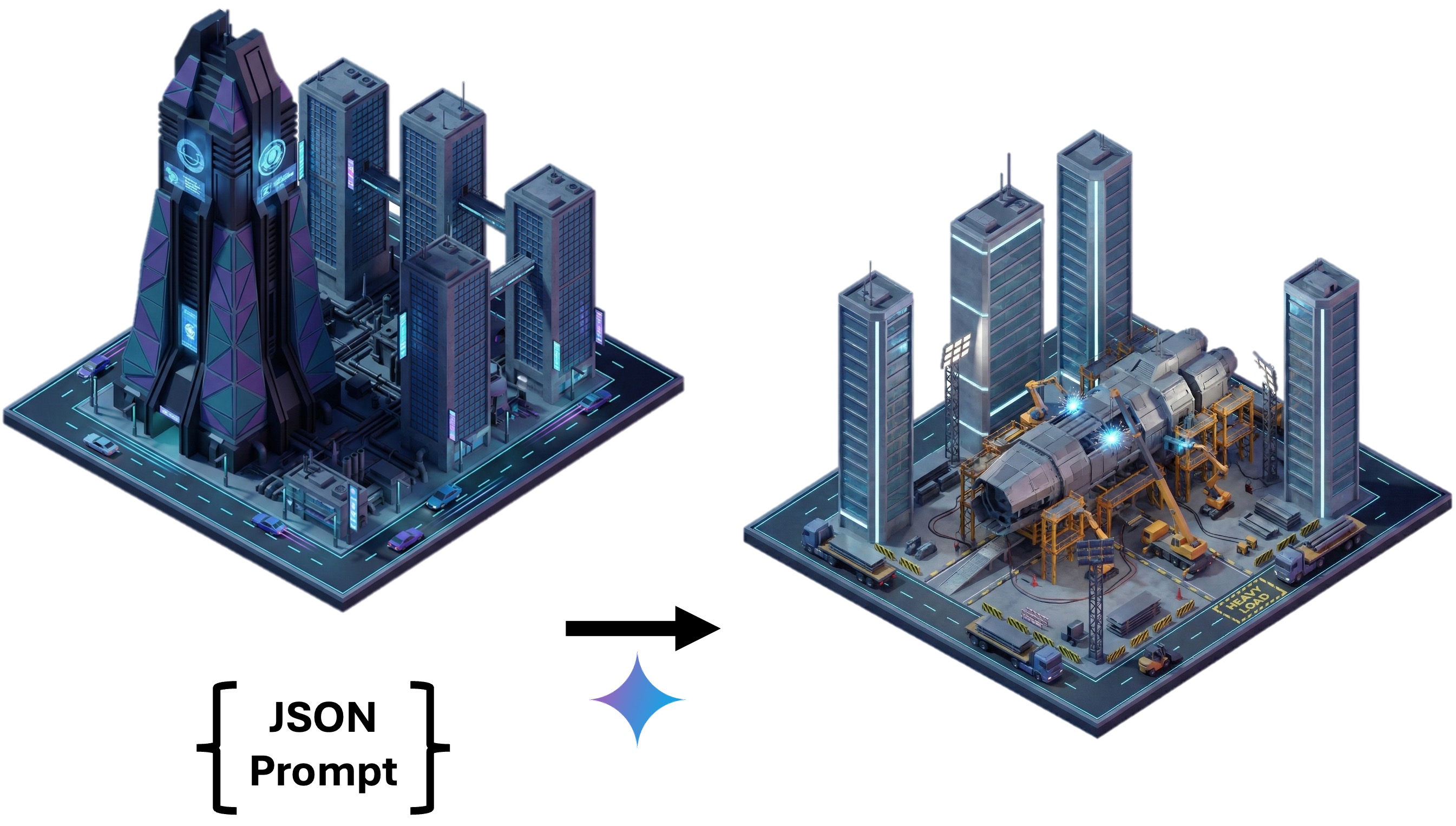}
\vspace{-20pt}
\caption{Prompting strategy to use image input as reference for style and accompanying JSON prompt for the content of the isometric chunk image.}
\label{fig:prompt_strategy2}
\end{figure}

The second strategy uses an additional image to specify the style of the generation. With JSON prompt to specify the actual contents of the image. We show an example of the second strategy in ~\cref{fig:prompt_strategy2} and corresponding JSON prompt in~\cref{lst:prompt2}. We note that this process is very finicky. With Nano Banana often ignoring instructions, and multiple re-writes of the JSON prompts would be needed by conversing with Gemini 3~\cite{comanici2025gemini, gemini3_2025}.

\begin{lstlisting}[language=json, caption={Prompt Strategy 2}, label={lst:prompt2}]
{
  "scene_metadata": {
    "style": "Low-poly 3D render, miniature diorama",
    "perspective": "45° top-down isometric, wide view with margins",
    "composition": "A self-contained, finished diorama chunk on a rectangular base, fully visible.",
    "background": "Solid white void"
  },
  "prompt_type": "scene_description",
  "art_style": {
    "view": "isometric",
    "angle": "45_degree_high_angle",
    "aesthetic": "low_poly_3D_render",
    "theme": "cyberpunk_shipyard_construction_site",
    "rendering": "smooth_shading, ambient_occlusion, industrial_floodlights_and_welding_arcs",
    "background": "Solid white with no shadow effects",
    "scene_visibility": "Entire scene is visible on a floating square landmass"
  },
  "elements": {
    "structures": [
      {
        "type": "spaceship_under_construction",
        "count": 1,
        "material": "unpainted_grey_metal_hull_segments, orange_and_yellow_scaffolding_beams, glowing_blue_welding_sparks",
        "details": "A massive, partially assembled cyberpunk spacecraft hull sits in the center. It is encased in a dense, complex lattice structure of low-poly scaffolding, work platforms, and ladders. Robotic arms with welding torches are actively working on the hull, creating bright flashes.",
        "layout": "Dominating the exact center of the diorama block, rising high above the ground level."
      },
      {
        "type": "research_and_engineering_skyscrapers",
        "count": 4,
        "material": "sleek_glass_panels, brushed_aluminum_frames, cool_blue_and_white_neon_strips",
        "details": "Tall, clean-looking high-rise buildings housing design labs and engineering firms. . They contrast with the gritty construction site.",
        "layout": "Positioned around the central construction site, overlooking the spaceship project."
      }
    ],
    "roads": {
      "type": "perimeter_loop_and_site_access",
      "style": "paved_asphalt_with_heavy_load_markings",
      "material": "dark_grey_asphalt, glowing_yellow_caution_stripes",
      "layout": "A continuous perimeter road enclosing the entire square base on all four sides. Internal temporary access roads made of reinforced concrete slabs lead from the perimeter to the central construction zone."
    },
    "vehicles": {
      "type": "heavy_construction_and_transport",
      "count": 8,
      "style": "low-poly_industrial_machinery",
      "details": "Large flatbed trucks carrying metal plates, mobile cranes lifting hull segments, and small robotic loaders moving materials.",
      "layout": "Driving along the perimeter roads and actively working within the central construction site."
    },
    "street_furniture": {
      "items": {
        "type": "construction_site_elements",
        "style": "utilitarian_and_temporary",
        "placement": "Tall temporary floodlight towers surrounding the spaceship, stacks of raw materials, construction barriers, and small automated worker bots on the scaffolds."
      }
    },
    "terrain": {
      "surface": "industrial_concrete_slab",
      "details": "The central ground level is heavy-duty concrete, marked with tire treads, oil stains, and temporary power cables running to the scaffolding.",
      "margins": "clean_cut_asphalt_edges_of_the
      _perimeter_road_floating_in_void",
    }
  },
  "lighting_and_atmosphere": {
    "time_of_day": "night",
    "shadows": "harsh, dynamic_shadows_from_floodlights_and_welding
    _flashes",
    "mood": "industrial, busy, constructive, high-tech_engineering, impressive scale",
  }
} Use reference image's style and content from the json prompt.
\end{lstlisting}

The rationale for both of these strategies is that we wanted generated images to have similar styles and settings such that NuiScene would have larger probability of piecing together different scenes during data generation. And that the elements across different scenes would appear natural when placed together.

\subsubsection{Colored Point Cloud Sampling} We mostly follow NuiScene's protocol for converting scenes into occupancies and sampling point clouds. We encourage reader's to see their paper for more details regarding the conversion process. To support the color training of our model, after we obtain the uncolored mesh from marching cubes of the converted occupancies from scenes. We first sample a point cloud along the surface of the output glb file from Trellis 2. We then use barycentric coordinates for the point cloud from their corresponding triangle face to convert to uvs and obtain RGB colors. Then for each vertex in the marching cubes mesh we assign it the color of its nearest neighbor. Finally, when we sample point clouds from the colored marching cubes mesh we can get both coordinates and color.

\section{Model Hyperparameters}
\label{sec:appendix_hyperparameter}

\subsection{NuiScene}

\subsubsection{VAE} Similar to NuiScene we use a sampled colored point cloud size of $4096$ as input. And the training is supervised with $4096$ query points for occupancy. In addition, we use sample another $2048$ points from the scene chunk and supervise the color training using coordinates as query and color as supervision for the prediction. The decoder consists of $24$ self-attention layers. The vector set size and usage of up-sampling layers is as specified in the main paper.

Training a VAE from scratch for each scenario is prohibitively time-consuming. Since we had already conducted experiments using two different sets of medieval images and bootstrapped scenes, we reuse the pretrained weights of this model and fine-tune it for the final medieval, desert, and cyberpunk scenarios. The pretrained model was initially trained on a set of bootstrapped medieval scenes for 160 epochs using AdamW~\cite{loshchilov2017decoupled} with a learning rate of 5e-5 and cosine annealing~\cite{loshchilov2016sgdr}. It was then further trained on a different set of medieval scenes using the same learning rate for an additional 60 epochs. For each final scenario, the model is fine-tuned separately using this set of weights for 60 epochs each. We conduct training on either 2 L40s GPUs or 4 A5000 GPUs with the effective total batch size of 40 quad chunks.

\subsubsection{Quad Chunk Diffusion} We mostly follow the same hyperparameters for training the Quad Chunk Diffusion model as in NuiScene. The main difference is that we switched from DDPM~\cite{ho2020denoising} to rectified flow~\cite{lipman2022flow}. Similar to SD3~\cite{esser2024scaling} we use logit normal sampling for time steps with mean=0 and std=1.

\subsection{Sketch to World Model}

For our default model, we use the sketch-to-world model with $512$ input channels for the vector set size of $(8, 64)$. With a model depth of $16$ and width of $1536$ and a total batch size of $24$. We train the model for $320$ epochs using AdamW with a learning rate of 1e-4 with cosine annealing. For rectified flow we use mean=1.0 and std=1.0 following the settings in Trellis~\cite{xiang2025structured}. The overfitting and full set training models were trained with 2 L40s GPUs. For the further fine-tuning with scenes sizes of $[1024, 1600]$, we train with 4 L40s GPUs with the same batch size $24$ and learning rate of 5e-5 for $320$ epochs.

\end{document}